\documentclass[letterpaper]{article}
\usepackage[preprint]{aaai2027}
\usepackage[hyphens]{url}
\usepackage{graphicx}
\usepackage{natbib}
\usepackage{caption}
\usepackage{amsmath,amssymb}
\usepackage{booktabs}
\usepackage{algorithm}
\usepackage{algorithmic}
\newcommand{\pos}[1]{[ #1 ]_+}
\newtheorem{proposition}{Proposition}

\title{HetGPS: Scalable Graph Multi-Agent Reinforcement Learning with Physics-Anchored Adaptive Safety for EV Charging}
\author{
Xiangwei Wang\textsuperscript{\rm 1},
Nanduni Nimalsiri\textsuperscript{\rm 2},
Yu Xia\textsuperscript{\rm 1},
Peng Wang\textsuperscript{\rm 3},
Saman Halgamuge\textsuperscript{\rm 1}
}
\affiliations{
\textsuperscript{\rm 1}The University of Melbourne, Australia\\
\textsuperscript{\rm 2}CSIRO, Australia\\
\textsuperscript{\rm 3}Shanghai Jiao Tong University, China
}

\begin{document}
\maketitle

\begin{abstract}
Safety interventions for large populations of network-coupled agents must
protect shared constraints without unnecessarily overriding task-oriented
policy decisions.  We present HetGPS, a hybrid graph-control framework
synergizing learned graph risk with physics-anchored correction by separating
intervention magnitude from corrective direction.  An action-conditioned graph
residual model schedules state-dependent intervention authority, while a
physics model determines its direction.  For electric vehicle (EV) charging,
we couple this filter with a parameter-shared heterogeneous-graph soft
actor-critic policy, enabling topology-aware coordination with a learned model
size independent of fleet size.  Across five nested distribution networks with
200--3,218 EVs and 100 evaluation days, Adaptive Authority reduces bus--step
voltage violations from 3.93--7.74\% without filtering to 0.52--3.44\%, while
maintaining 99.06--100\% departure success.  Relative to the same
physics-directed projection with fixed authority, it improves mean reward on
all five networks and lowers the mean safety score on four.  The deployed
policy-and-risk model contains 383,702 learned parameters at every scale; at
3,218 EVs, a matched centralized SAC actor is about $170\times$ larger.  A
policy trained on the eight-transformer system transfers zero-shot to the
16- and 32-transformer systems, attaining 0.57--0.75\% violation rates and at
least 99.99\% departure success.  These results show that learned graph risk
can allocate intervention authority at scale while feeder physics anchors
corrective action.
\end{abstract}

\section{Introduction}

Coordinating residential EV charging presents interconnected challenges in scalability and network safety.  Many locally controlled EV chargers have heterogeneous private
states, yet their actions remain coupled through distribution-grid voltages
~\cite{ma_decentralized_2013,yan_cooperative_2022}.

\begin{figure}[!t]
\centering
\includegraphics[width=\columnwidth]{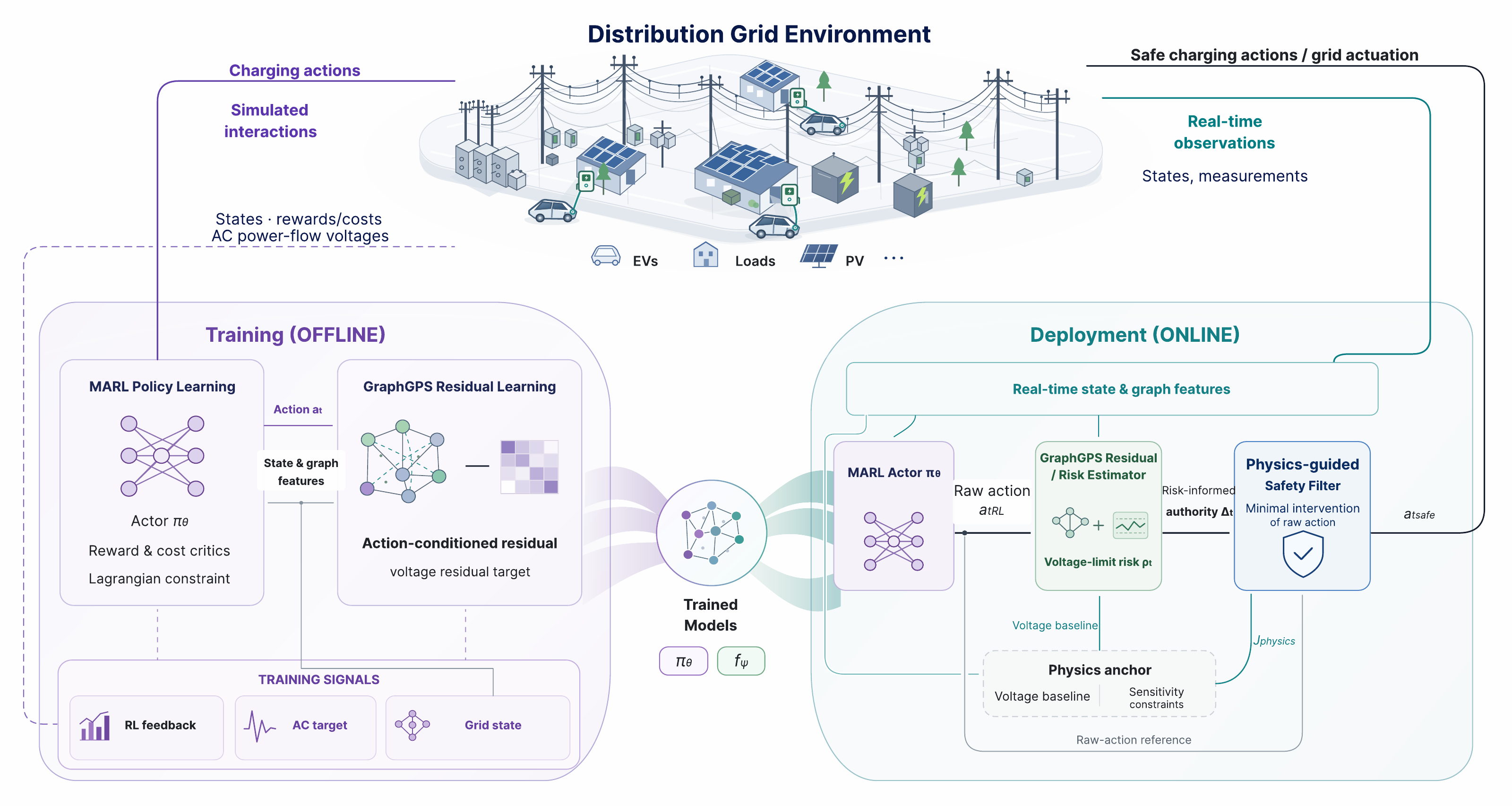}
\caption{End-to-end EV-control pipeline.  Offline, the shared MARL policy and
action-conditioned graph residual model learn from power-flow interactions.
At deployment, the policy proposes joint EV actions; before grid actuation, the
safety filter evaluates their predicted voltage consequences and corrects them
when needed.  Learned residual risk sets the maximum correction magnitude
$\Delta_t$, while feeder physics determines its direction.}
\label{fig:framework}
\end{figure}

Uncoordinated charging overlooks the electrical interactions that couple the charging decisions of multiple EV agents. In contrast, centralized coordination becomes increasingly complex as the number of EVs and the size of the network grow. Graph-based parameter sharing offers a scalable middle ground
by applying a shared policy to topology-aware local information
~\cite{rampasek_recipe_2022,yan_multi-agent_2024}.  Scalability alone, however,
does not make the resulting actions safe.

\newpage

Training-time reward penalties or expected-cost constraints can reduce
violations on average, but they do not ensure that every realized joint
charging action is acceptable under changing demand, PV generation, and EV
availability.  A predictive safety filter is a runtime layer placed between a
learned policy and the physical system.  Before an action is executed, it uses
current measurements and a model to predict the action's constraint
consequences; it leaves the proposal unchanged when no correction is needed and
otherwise returns a nearby corrected action~\cite{wabersich_predictive_2021}.
Here, the predicted quantity is the post-action bus-voltage profile.  The filter
therefore protects network voltage while the MARL policy continues to optimize
charging cost and departure readiness~\cite{li_constrained_2020}.

The remaining design question is how much correction the filter should be
allowed to apply.  A weak fixed limit may be insufficient under high voltage
risk, whereas a worst-case limit can unnecessarily distort useful actions in
benign conditions.  Rather than asking one learned model to infer both this
magnitude and the physically appropriate direction, we separate the two roles.
We propose HetGPS, a scalable graph MARL framework with an adaptive-authority
safety filter, summarized in Figure~\ref{fig:framework}.  A graph-based risk
model schedules a state-dependent authority $\Delta_t$, defined as the maximum
permitted change to the proposed EV actions, while a feeder-physics model
determines the voltage-corrective direction.  Thus, learning decides how
strongly the filter may intervene, whereas physics decides how the intervention
should modify the actions.

The EV instantiation uses a heterogeneous bus--agent graph and a shared
GraphGPS-derived policy~\cite{rampasek_recipe_2022}.  Bus message passing exposes
each charger to topology-aware electrical context without pooling every
household's private EV state in a fleet-sized centralized critic.  An
action-conditioned voltage residual head is supervised by power-flow labels
during training and reused as the deployment risk model.  HetGPS reuses the
same graph-encoder and residual-risk-head weights across buses, and the same
soft actor-critic (SAC) policy weights across
EVs~\cite{haarnoja_soft_2018}.  Expanding the graph therefore applies the same
learned modules to more nodes and agents rather than creating new parameters.
It increases computation and topology-dependent physical
buffers, while the deployed policy-and-risk model remains at 383,702 learned
parameters across all evaluated scales.  For scale, the largest tested CRE21
instance contains 32 distribution transformers and 3,218
EVs~\cite{team_nando_mvlv_2023}.  On this instance, a fully centralized SAC
policy that
concatenates all EV observations and jointly outputs all actions contains
65.2 million parameters, about $170\times$ as many.  Matched deployment- and
training-model accounting is reported in Appendix~\ref{app:secondary}.  To
further assess generalization, we keep the learned weights fixed, rebuild the
graph and physical-sensitivity buffers for each target feeder, and evaluate the
controller on unseen topologies.

Our contributions are:
\begin{itemize}
    \item We introduce a hybrid safety-filter factorization in which a learned
    graph risk model schedules state-dependent intervention authority, while a
    physics model determines the corrective direction.
    \item We instantiate this principle as HetGPS, combining a
    parameter-shared graph MARL policy, an action-conditioned voltage-residual
    model, and a physics-anchored projection.  Its learned parameter count
    remains fixed across fleet sizes.  During operator-mediated execution,
    household-level EV states remain local: each agent receives topology-aware
    bus context without direct access to neighboring agents' private features.
    \item We demonstrate the resulting reward--safety trade-off on
    distribution networks with 200--3,218 EVs through controlled no-filter,
    fixed-authority, and adaptive-authority comparisons, and further evaluate
    zero- and few-shot transfer to larger and cross-family network topologies.
\end{itemize}

\section{Related Work}

\textbf{Safe and constrained RL.}
Constrained policy optimization and decentralized primal--dual safe RL regulate
expected cumulative costs
~\cite{achiam_constrained_2017,lu_decentralized_2021,stooke_responsive_2020},
whereas shields, learned editors, and predictive filters intervene on individual
actions~\cite{alshiekh_shielding_2018,yu_safety_editor_2022,
wabersich_predictive_2021}.  Adaptive regularization can vary reliance on a
hard-coded safety component~\cite{tian_adaptive_2024}.  HetGPS instead adapts
only deployment-time intervention authority; feeder physics retains the
correction direction.

\textbf{Graph learning for networked control.}
Graph policies exploit relational structure for control and dynamic cooperation
~\cite{wang_nervenet_2018,jiang_graph_convolutional_2020}; selective aggregation
and factor graphs address large populations~\cite{fu_concentration_2022,
fan_efficient_collaboration_2025}.  Power-system graph models learn optimal
power flow or voltage control~\cite{owerko_optimal_2020,
cao_physics-informed_2024,yan_multi-agent_2024}, while recent AI work poses
active voltage control as Dec-POMDP MARL or a constrained Markov game
~\cite{wang_active_voltage_2021,qu_safety_constrained_2024}.  HetGPS separates
graph coordination from action-conditioned risk estimation and tests shared
weights directly on unseen scales and feeder families.

\textbf{EV coordination.}
AI work spans distributed smart charging, multi-agent EV benchmarks under
distribution shift, and optimization-guided heterogeneous vehicle-to-building
control~\cite{valogianni_smart_2014,yeh_sustaingym_2023,liu_v2b_2025}.
Power-system studies treat price-responsive scheduling and constrained service
objectives~\cite{ma_decentralized_2013,silva_coordination_2020,
wan_model-free_2019,li_constrained_2020}.  Distributed and privacy-aware
controllers reduce information pooling~\cite{liu_decentralized_2019,
nimalsiri_distributed_2024,qin_privacy_2021}.  HetGPS adds topology-scalable
coordination and deployment-time physics-anchored voltage correction.  Each
actor uses local EV state and operator-derived bus context, not neighboring
EVs' raw features; this is information locality, not formal privacy.

\section{Methodology}

\subsection{Problem Setting}

We model the graph-coupled cooperative system as a constrained Dec-POMDP over
an agent set $\mathcal A$ of $N_a=|\mathcal A|$ agents, with state $x_t$,
transition kernel $P$, reward and cost discounts $\gamma,\gamma_c$, and a
finite horizon of $T=24$ hourly decisions.  Observations are a deterministic
function of the state, $o_i^t=h_i(x_t)$.
Let $\chi_i^t$ denote agent $i$'s private observation component.  An
operator-side graph encoder aggregates serving-bus, neighboring-bus, and
feeder-level global context from feeder topology and bus-level measurements
into an agent-specific context $c_i(\mathcal G_t)$ that carries no neighboring
EV's raw private state.  The complete policy observation is
$o_i^t=[\chi_i^t\|c_i(\mathcal G_t)]$, and every executed action passes through
three stages:
\begin{equation}
\begin{aligned}
 a_{i,t}^{\mathrm{RL}}&\sim\pi_\theta(\cdot\mid o_i^t)
   &&\text{learned proposal},\\
 a_t^{\mathrm{proj}}&=F_{\Delta_t}(a_t^{\mathrm{RL}})
   &&\text{voltage projection},\\
 a_t^{\mathrm{exec}}&=G_{\mathrm{svc}}(x_t,a_t^{\mathrm{proj}})
   &&\text{service map},
\end{aligned}
 \label{eq:policy}
\end{equation}
with $a_t^{\mathrm{RL}}=(a_{i,t}^{\mathrm{RL}})_{i\in\mathcal A}$.  The
projection $F_{\Delta_t}$ is the safety filter studied here, bounded by the
authority $\Delta_t$; the service map $G_{\mathrm{svc}}$ deterministically
enforces availability, SoC, and departure rules and is always active.  Only
$F_{\Delta_t}$ is optional: policy training sets it to the identity, so
training optimizes $\Pi_\theta^{\mathrm{tr}}=G_{\mathrm{svc}}\circ\pi_\theta$
while deployment runs $\Pi_\theta^{\mathrm{dep}}
=G_{\mathrm{svc}}\circ F_{\Delta_t}\circ\pi_\theta$.  The normative
system-level criterion for an executed controller $\Pi_\theta$ is
\begin{equation}
\begin{aligned}
 \max_\theta\quad J_R(\Pi_\theta)
   &=\mathbb E_{\Pi_\theta,P}\!\left[\sum_{t=0}^{T-1}\gamma^t R(x_t,a_t)\right],\\
 \text{s.t.}\quad J_{C,b}(\Pi_\theta)
   &=\mathbb E_{\Pi_\theta,P}\!\left[\sum_{t=0}^{T-1}\gamma_c^t C_b(x_t,a_t)\right]
     \leq\epsilon_b,\quad \forall b\in\mathcal B_{\mathrm{mon}}.
\end{aligned}
\label{eq:cmdp_objective}
\end{equation}
Here $a_t=a_t^{\mathrm{exec}}$ inside $R$ and $C_b$, which is a nonnegative
post-action AC-voltage cost at monitored bus $b$ with prescribed expected
cumulative budget $\epsilon_b$.  The shared critics introduced below approximate
this expectation-constrained objective for
$\Pi_\theta^{\mathrm{tr}}$; the deployment projection is a post-training action
interface rather than part of their Bellman targets.  The objective does not
assert pointwise feasibility for every realization.  The implementation uses
$\gamma_c=\gamma$ for Bellman training, whereas all reported voltage metrics
and deployment decisions use undiscounted stepwise AC voltages.
Replay stores the raw SAC proposal, while rewards and post-action AC-voltage
labels are generated after $G_{\mathrm{svc}}$.  Deployment inserts the voltage
projection between the proposal and the same service map.

The executed EV action $a_{i,t}^{\mathrm{exec}}\in[-1,1]$ maps to
demand-positive charging ($a_{i,t}^{\mathrm{exec}}>0$) or V2G discharge
($a_{i,t}^{\mathrm{exec}}<0$), with presence, charger-rating, and
state-of-charge (SoC) limits enforced by the transition model.  The reward
combines real-time energy cost, throughput-based battery
wear~\cite{ortegavazquez_optimal_2014,farzin_practical_2016}, a continuous
readiness service cost, and departure shortfall.  The readiness term is part
of the optimized task reward, not potential-based policy-invariant shaping.
We assume a balanced positive-sequence network; pandapower AC power flow
evaluates the operating band
$[V^{\min},V^{\max}]=[0.95,1.05]$ p.u.~\cite{thurner_pandapoweropen-source_2018}.

The heterogeneous graph has bus nodes, EV nodes, physical line/transformer
edges, and EV-to-bus attachment edges.  Bus nodes expose measured voltage and
aggregate power features, while an EV's SoC, availability, target, departure,
and histories remain private EV features.  Attachment edges deliver the
serving-bus embedding to the shared actor.  EV dynamics, power flow, and full
feature definitions are given in Appendices~\ref{app:physical}
and~\ref{app:graph}.

\subsection{Graph-Guided Authority, Physics-Anchored Direction}

Figure~\ref{fig:framework} summarizes the offline-training and online-deployment
pipeline while separating policy, learned authority, and physical direction.
Here $y$ is the monitored voltage and $J_t$ is its nominal physical
sensitivity to the EV actions.

\subsubsection{Shared Graph Policy and Risk Model.}

Two bus message-passing layers encode measured electrical state and feeder edge
attributes into operator-side embeddings.  The graph encoder compresses the
variable-size, operator-visible electrical state into a fixed-dimensional
context for each agent, avoiding fleet-wide state concatenation and thereby
mitigating the associated curse of dimensionality in centralized control.  One
Gaussian SAC actor is shared
across agents~\cite{haarnoja_soft_2018}; different observations still yield
individual actions.  Shared mean-field reward and voltage-cost critics
condition additionally on the mean and standard deviation of the raw fleet
action proposals.  Together with PID-updated bus multipliers, they provide a
scalable approximation to Eq.~\eqref{eq:cmdp_objective}
\cite{stooke_responsive_2020}.

For each bus $v$, the action-conditioned residual head takes the raw policy
proposal as input and is supervised to predict the safety output after the
environment's common deterministic service transformation:
\begin{equation}
 \widetilde y^{(\tau)}_{t,v}(a_t^{\mathrm{RL}})=y^{\mathrm{fb}}_{t,v}
 +[J_t a_t^{\mathrm{RL}}]_v
 +\widehat r^{(\tau)}_{\phi,v}(\mathcal G_t,a_t^{\mathrm{RL}}),
 \label{eq:residual}
\end{equation}
Here, $y_t^{\mathrm{fb}}=V_{t-1}^{\mathrm{obs}}$ is the latest pre-decision
voltage measurement, and $J_t$ is the nominal physics sensitivity derived from
LinDistFlow, a linearized radial power-flow model that maps nodal power
injections to approximate voltage changes~\cite{baran_network_1989}.  AC
power-flow labels supervise the residual head.  The joint action is not
concatenated into a fleet-sized dense input: a shared encoder maps each
agent embedding and proposal, and these representations are sum-pooled at the
serving buses before shared bus processing.  A pinball loss trains a median
head together with a lower/upper pair that defines an uncalibrated interval; we
write $\widetilde y_{t,v}$ for the median used below.  The three quantile
outputs share one trunk and are counted together in the deployed parameter
total, although deployment evaluates only the median.  Authority scheduling evaluates that
median at $a_t^{\mathrm{RL}}$; the separate calibration audit re-evaluates all
three heads with $a_t^{\mathrm{exec}}$ as their input.  The head neither
predicts $\Delta_t$ directly nor supplies the projection direction.

\subsubsection{Risk-to-Authority Scheduler.}

Let $b_i$ be the serving bus of agent $i$, and for a predicted voltage field
$y$ let $z_v(y)=\pos{V^{\min}+\delta_m-y_v}+\pos{y_v-V^{\max}+\delta_m}$ be its
exceedance of the operating band at bus $v$, tightened by an internal margin
$\delta_m$.  We evaluate this exceedance both on the learned median
$\widetilde y_t$ and on a fixed-gain physics guard
$y^{\mathrm{guard}}_t=y_t^{\mathrm{fb}}+\beta J_t a_t^{\mathrm{RL}}$, aggregate
each over agents by a high quantile, and keep the more conservative value:
\begin{equation}
 \bar r_t=\max\bigl\{
 Q_{.95}\{z_{b_i}(\widetilde y_t):i\in\mathcal A\},\;
 Q_{.95}\{z_{b_i}(y^{\mathrm{guard}}_t):i\in\mathcal A\}\bigr\}.
 \label{eq:risk}
\end{equation}
Calibrating $\beta$ against the measured AC sensitivity yields $\beta<1$, so
the guard attenuates rather than amplifies; LinDistFlow overstates action
sensitivity on these feeders.  The quantile runs over agent--bus incidences,
which coincide with distinct buses whenever the serving map is injective, as it
is throughout CRE21.  This risk sets the authority but does not decide when the
filter acts.  A monotone map turns it into the intervention budget:
\begin{equation}
 \Delta_t=\Delta_{\min}+(\Delta_{\max}-\Delta_{\min})
 \operatorname{clip}\!\left(
 \frac{\bar r_t-r_{\mathrm{low}}}{r_{\mathrm{high}}-r_{\mathrm{low}}},0,1
 \right).
 \label{eq:authority}
\end{equation}
The filter is triggered when any physics-guard incidence lies outside the
internally margined operating band; the unscaled physical sensitivity $J_t$
then defines the correction below.  Detection and budgeting use different
aggregators by design: detection must react to any exceedance, whereas the
budget should follow the magnitude of widespread risk rather than a single
outlier.  This does not collapse the schedule onto $\Delta_{\min}$, because the
learned exceedance field is dense rather than sparse when the feeder is
loaded; on our traces the scheduled authority never reaches its floor on a
triggered step.
The monitored set is the distinct EV-hosting buses, and the deepest excursion
is never observed at a non-monitored bus, whose violation rate the filter also
reduces.

\subsubsection{Physics-Anchored Projection.}

The projection deliberately excludes the learned residual from its correction
direction, predicting voltage from physics alone as
$\widehat y_t^{\mathrm{phys}}(a)=y_t^{\mathrm{fb}}+J_t a$.  When the guard
flags a set $\mathcal K_t$ of buses, the smallest correction
$d=a-a_t^{\mathrm{RL}}$ that restores them is the convex quadratic program
\begin{equation}
\begin{aligned}
 \min_{d}\quad&\tfrac12\|d\|_W^2\\
 \text{s.t.}\quad&c_j^\top d\geq h_j,&&j\in\mathcal K_t,\\
 &\ell_{i,t}\leq d_i\leq u_{i,t},&&i\in\mathcal A,
\end{aligned}
\label{eq:qp}
\end{equation}
where $W$ is a fixed positive definite metric, $c_j$ is the row of $J_t$ at the
flagged bus signed by the violated side, $h_j$ is its band deficit, and
$\ell_{i,t}=\max\{-1-a_{i,t}^{\mathrm{RL}},-\Delta_t\}$,
$u_{i,t}=\min\{1-a_{i,t}^{\mathrm{RL}},\Delta_t\}$ form the authority box
through which the scheduled budget enters.  Solving Eq.~\eqref{eq:qp} exactly
every hour for thousands of agents is too costly, so the deployed controller
applies a fixed number of projection-onto-convex-sets sweeps: each flagged
constraint is relaxed in turn along its own $c_j$, in decreasing order of
deficit, and the iterate is projected back onto the box.  Every correction
direction is therefore physical, while the learned model only sizes the box.
The tested learned-direction alternatives produced no consistent Pareto
improvement; equal-cap diagnostics are reported in
Appendix~\ref{app:learned_direction}.
Because the sweeps are truncated, the result respects the action and authority
limits but need not satisfy every voltage constraint.

The service map then has the last word.  It zeroes an unavailable EV whatever
the projection commanded, so part of the authority budget can be spent on
agents that $G_{\mathrm{svc}}$ removes, and departure overrides can reverse a
correction outright.  Guarantees are therefore stated for the executed action
rather than the projected one.

\begin{algorithm}[t]
\caption{Graph-Guided Authority with Physics-Anchored Direction}
\label{alg:authority_filter}
\small
\begin{algorithmic}[1]
\REQUIRE Graph state $\mathcal G_t$, policy observations $\{o_i^t\}$,
feedback $y_t^{\mathrm{fb}}$, sensitivity $J_t$, operating limits
\STATE Sample $a_t^{\mathrm{RL}}$ from the shared graph policy
\STATE Predict $\widetilde y_t(a_t^{\mathrm{RL}})$ using
Eq.~\eqref{eq:residual}
\STATE Compute learned and guard risks using Eq.~\eqref{eq:risk}
\STATE Schedule authority $\Delta_t$ using Eq.~\eqref{eq:authority}
\STATE Form active constraint set $\mathcal K_t$ from the physics guard
\IF{no guard incidence is active}
  \STATE $a_t^{\mathrm{proj}}\gets a_t^{\mathrm{RL}}$
\ELSE
  \STATE Initialize $d\gets0$ and the authority box $[\ell_t,u_t]$
  \FOR{each projection sweep and $j\in\mathcal K_t$}
    \STATE Relax constraint $j$, then reproject $d$ onto $[\ell_t,u_t]$
  \ENDFOR
  \STATE $a_t^{\mathrm{proj}}\gets
  \operatorname{clip}(a_t^{\mathrm{RL}}+d,-1,1)$
\ENDIF
\STATE $a_t^{\mathrm{exec}}\gets
G_{\mathrm{svc}}(x_t,a_t^{\mathrm{proj}})$
\RETURN $a_t^{\mathrm{exec}}$
\end{algorithmic}
\end{algorithm}

Algorithm~\ref{alg:authority_filter} makes the intervention factorization
explicit.  The learned path is evaluated once to set a scalar budget; it never
changes the physical sensitivity or the constraint normals.  For
$K_{\mathrm{proj}}$ projection sweeps and $|\mathcal K_t|$ active one-sided
constraints, the dense projection cost is
$O(K_{\mathrm{proj}}|\mathcal K_t|N_a)$, while the learned parameter count is
independent of $N_a$.

\section{Experiments}

\subsection{Networks, Data, and Splits}

The primary benchmark uses five nested selections from CRE21, an open model of
a real Australian urban distribution feeder with synthesised LV
networks~\cite{team_nando_mvlv_2023}, containing 200, 664, 1236, 2239,
and 3,218 EVs (K=1,4,8,16,32).  Every household hosts an EV, so charging demand
dominates the residential base load and drives the voltage excursions we study.
IEEE 13/33/69 feeders provide secondary cross-family tests.  Feeder settings
are listed in Appendix~\ref{app:secondary}.

Each episode contains 24 hourly decisions.  Demand/PV profiles come from the
2012--2013 Ausgrid data~\cite{ratnam_residential_2017}; the price signal is the
absolute value of hourly averaged 2022--2023 AEMO VIC1 real-time wholesale
price~\cite{aemo_price_demand_2022}; EV mobility follows VISTA-derived
distributions~\cite{vista_2020}.  Chargers are heterogeneous in battery
capacity and rating; the fleet composition and profile preprocessing are
detailed in Appendix~\ref{app:reproducibility}.  Of the 365 days, 240 are used for training, 100
for evaluation, and 25 for topology transfer adaptation.  All evaluation arms see the
same days in the same order, so comparisons are paired.  All experiments are
implemented in PyTorch, and every reported result is produced on a single
NVIDIA A100 80GB GPU.

\subsection{Controlled Comparisons and Metrics}

We run two comparisons, each varying one component while holding the other
fixed.  The controlled backbone comparison fixes the safety interface across
HetGPS, ISAC, centralized SAC~\cite{haarnoja_soft_2018}, and
MADDPG~\cite{lowe_multi-agent_nodate}.  We additionally report PI-LP, a
non-learning perfect-information reference that schedules bidirectional EV
power over the full horizon using a linearized network model.

The second fixes the policy and varies the safety interface: (i) \emph{No
Filter}; (ii) \emph{Fixed Authority}, the physics projection with
$\Delta_t\equiv\Delta_{\max}$; and (iii) \emph{Adaptive Authority}, the same
projection with $\Delta_t$ from Eq.~\eqref{eq:authority}.
All three arms retain the same post-projection service guard; ``No Filter''
therefore means no \emph{voltage} projection, not the removal of physical
availability or departure-feasibility logic.  All metrics are computed on the
executed action.

The unified filter configuration is
$(\Delta_{\min},\Delta_{\max})=(0.20,0.35)$,
$(r_{\mathrm{low}},r_{\mathrm{high}})=(0.005,0.080)$ p.u., guard gain
$\beta=0.31$, internal margin $\delta_m=0$, quantile levels
$\tau\in\{0.05,0.50,0.95\}$, Euclidean metric $W=I$, and
$K_{\mathrm{proj}}=10$ projection sweeps.  The residual head's interval outputs
characterize predictive risk; they do not certify the deployed controller.

For episode $e$, step $t$, and monitored EV-hosting bus $v$, define
\begin{equation}
 d_{e,t,v}=\pos{V^{\min}-V_{e,t,v}}+\pos{V_{e,t,v}-V^{\max}}.
\end{equation}
The bus--step violation rate is the total count of $d_{e,t,v}>0$ divided by
$\sum_e T_e|\mathcal B_{\mathrm{EV},e}|$; distinct EV-hosting buses, not all graph
buses or EV agents, form the denominator.  The episode safety score is
\begin{equation}
 m_{s,e}=\frac{1}{T_e}\sum_t\mathbf 1\{\max_v d_{e,t,v}>0\}
 +\frac{\max_{t,v}d_{e,t,v}}{V^{\max}-V^{\min}},
 \label{eq:ms}
\end{equation}
and we report its mean over days.  Reward per EV and departure-target success
measure utility and service; a departure is counted successful when its SoC is
within an absolute tolerance of 0.10 below the requested target.  Lower
violation rate and $m_s$ are safer; higher
(less negative) reward is better.

\textbf{Topology-transfer protocol.}
We evaluate a K=8-trained controller zero-shot on K=16, K=32, and IEEE33, with
a 25-episode K=32 adaptation arm.  K16 and K32 retain the source feeder
settings, whereas IEEE33 tests a topology and operating-distribution shift.  Target
electrical buffers are rebuilt from each feeder, while learned shared weights
are copied without target training in the zero-shot arm.

\section{Results}

\subsection{Policy Training Dynamics}

\begin{figure}[t]
\centering
\makebox[\columnwidth][c]{%
\includegraphics[width=1.06\columnwidth]{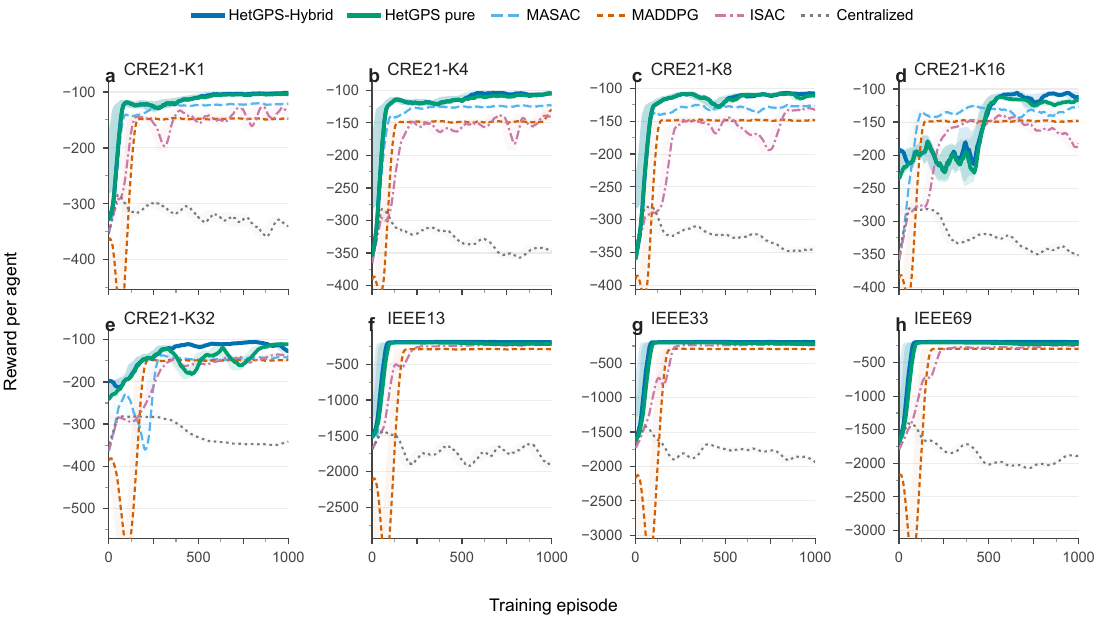}}
\caption{Policy-training reward per agent (EV) over 1,000 episodes for the five
nested CRE21 systems and three IEEE feeders.}
\label{fig:training_curves}
\end{figure}

Figure~\ref{fig:training_curves} shows optimization behaviour across all eight
evaluated network scenarios.  The HetGPS backbone converges faster than the
baselines in most scenarios and reaches a strong final reward across all eight.
Checkpoints are selected on service quality rather than by taking the final
iterate.  Table~\ref{tab:backbone} compares the four learned backbones and
PI-LP on all eight scenarios without a voltage filter.  HetGPS gives the
highest feasible reward on all five CRE21 scales and on IEEE33 and IEEE69, with
its largest advantage at K=32; MADDPG retains a small reward margin only on
IEEE13.  The
centralized controller's low voltage metrics coincide with only 72--80\%
departure success and therefore do not represent a usable operating point.
Complete numerical results, including departure success for every method and
scenario, are provided in Appendix~\ref{app:secondary}.

\begin{table}[!t]
\centering
\scriptsize
\setlength{\tabcolsep}{1.8pt}
\renewcommand{\arraystretch}{0.84}
\resizebox{\columnwidth}{!}{%
\begin{tabular}{@{}llrrr@{}}
\toprule
Scenario & Method & Reward/EV $\uparrow$ & Viol. (\%) $\downarrow$ & $m_s\downarrow$\\
\midrule
CRE21 $K{=}1$ & HetGPS & $\boldsymbol{-100.31\pm2.60}$ & $0.043\pm0.033$ & $0.015\pm0.010$\\
 & ISAC & $-104.42\pm1.08$ & $\boldsymbol{0.001\pm0.001}$ & $\boldsymbol{0.001\pm0.001}$\\
 & MADDPG & $-138.90\pm25.36$ & $0.103\pm0.002$ & $0.030\pm0.001$\\
 & C-SAC & $-353.67\pm17.96$ & $0.000\pm0.000$ & $0.000\pm0.000$\\
 & PI-LP & $-135.46$ & $1.409$ & $0.133$\\
\addlinespace[1pt]
CRE21 $K{=}4$ & HetGPS & $\boldsymbol{-103.51\pm1.20}$ & $0.058\pm0.012$ & $0.059\pm0.007$\\
 & ISAC & $-108.38\pm2.11$ & $\boldsymbol{0.017\pm0.002}$ & $\boldsymbol{0.019\pm0.002}$\\
 & MADDPG & $-115.04\pm4.21$ & $0.133\pm0.002$ & $0.105\pm0.001$\\
 & C-SAC & $-382.75\pm15.81$ & $0.000\pm0.000$ & $0.000\pm0.000$\\
 & PI-LP & $-136.31$ & $1.288$ & $0.184$\\
\addlinespace[1pt]
CRE21 $K{=}8$ & HetGPS & $\boldsymbol{-103.62\pm2.63}$ & $0.019\pm0.010$ & $0.033\pm0.014$\\
 & ISAC & $-121.89\pm3.14$ & $\boldsymbol{0.002\pm0.002}$ & $\boldsymbol{0.005\pm0.005}$\\
 & MADDPG & $-126.76\pm23.53$ & $0.083\pm0.005$ & $0.096\pm0.004$\\
 & C-SAC & $-383.91\pm2.29$ & $0.000\pm0.000$ & $0.000\pm0.000$\\
 & PI-LP & $-136.88$ & $1.171$ & $0.183$\\
\addlinespace[1pt]
CRE21 $K{=}16$ & HetGPS & $\boldsymbol{-107.22\pm5.43}$ & $0.007\pm0.012$ & $0.020\pm0.032$\\
 & ISAC & $-127.33\pm6.49$ & $\boldsymbol{0.005\pm0.006}$ & $\boldsymbol{0.016\pm0.020}$\\
 & MADDPG & $-136.31\pm30.35$ & $0.072\pm0.020$ & $0.109\pm0.015$\\
 & C-SAC & $-389.18\pm7.60$ & $0.000\pm0.000$ & $0.000\pm0.000$\\
 & PI-LP & $-136.85$ & $1.137$ & $0.204$\\
\addlinespace[1pt]
CRE21 $K{=}32$ & HetGPS & $\boldsymbol{-108.45\pm8.21}$ & $\boldsymbol{0.001\pm0.001}$ & $\boldsymbol{0.003\pm0.003}$\\
 & ISAC & $-126.78\pm4.97$ & $0.002\pm0.002$ & $0.012\pm0.014$\\
 & MADDPG & $-153.99\pm0.02$ & $0.061\pm0.001$ & $0.122\pm0.001$\\
 & C-SAC & $-389.56\pm4.51$ & $0.000\pm0.000$ & $0.000\pm0.000$\\
 & PI-LP & $-137.69$ & $1.010$ & $0.201$\\
\midrule
IEEE13 & HetGPS & $-190.65\pm0.01$ & $\boldsymbol{0.000\pm0.000}$ & $\boldsymbol{0.000\pm0.000}$\\
 & ISAC & $-213.91\pm5.11$ & $\boldsymbol{0.000\pm0.000}$ & $\boldsymbol{0.000\pm0.000}$\\
 & MADDPG & $\boldsymbol{-189.80\pm0.01}$ & $\boldsymbol{0.000\pm0.000}$ & $\boldsymbol{0.000\pm0.000}$\\
 & C-SAC & $-1874.91\pm128.20$ & $0.000\pm0.000$ & $0.000\pm0.000$\\
 & PI-LP & $-592.47$ & $0.125$ & $0.017$\\
\addlinespace[1pt]
IEEE33 & HetGPS & $\boldsymbol{-191.29\pm0.07}$ & $\boldsymbol{3.529\pm0.001}$ & $0.441\pm0.001$\\
 & ISAC & $-214.20\pm2.14$ & $3.646\pm0.019$ & $0.425\pm0.003$\\
 & MADDPG & $-192.06\pm0.04$ & $3.650\pm0.001$ & $0.441\pm0.001$\\
 & C-SAC & $-2014.02\pm75.71$ & $0.227\pm0.063$ & $0.039\pm0.011$\\
 & PI-LP & $-479.94$ & $3.814$ & $\boldsymbol{0.336}$\\
\addlinespace[1pt]
IEEE69 & HetGPS & $\boldsymbol{-188.75\pm0.01}$ & $\boldsymbol{0.320\pm0.001}$ & $0.065\pm0.001$\\
 & ISAC & $-219.68\pm2.81$ & $0.409\pm0.003$ & $\boldsymbol{0.050\pm0.001}$\\
 & MADDPG & $-189.63\pm0.07$ & $0.410\pm0.000$ & $0.065\pm0.001$\\
 & C-SAC & $-2029.39\pm43.36$ & $0.000\pm0.000$ & $0.000\pm0.000$\\
 & PI-LP & $-570.14$ & $2.101$ & $0.197$\\
\bottomrule
\end{tabular}%
}
\caption{Backbone comparison on all eight scenarios.  Learned entries are mean
$\pm$ sample standard deviation across three runs, each averaged over the same
100 evaluation days; PI-LP is deterministic.  Bold marks the best value in each
metric among controllers with at least 95\% departure success.  Full service
results are in Appendix~\ref{app:secondary}.}
\label{tab:backbone}
\end{table}

\subsection{Reward--Safety Trade-offs of Adaptive Authority}

\begin{figure}[!t]
\centering
\includegraphics[width=\columnwidth]{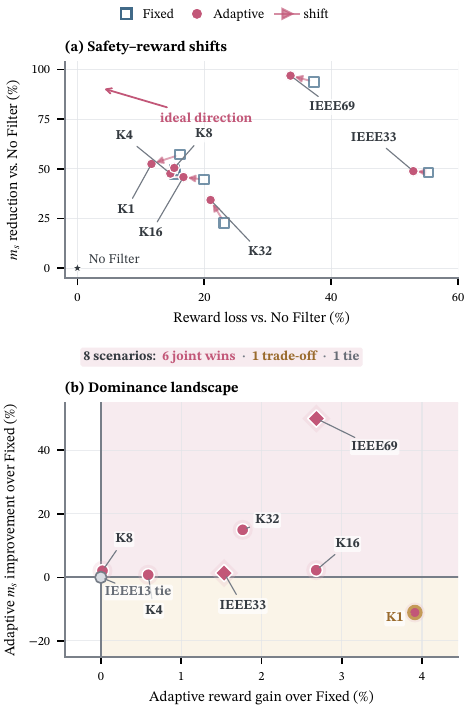}
\caption{Reward--safety comparison using the three-run means.  (a) Fixed and
adaptive authority relative to No Filter; each arrow points from Fixed to
Adaptive, with upward and leftward shifts indicating more $m_s$ reduction and
less reward loss.  (b) Adaptive improvement over Fixed; the upper-right region
indicates joint improvement.}
\label{fig:dominance}
\end{figure}

Figure~\ref{fig:dominance} separates the protection benefit of filtering from
the value of adapting its authority.  Relative to matched fixed authority,
adaptive authority improves reward and $m_s$ jointly on six of the seven
nontrivial networks.  K1 instead recovers reward with a modest safety
trade-off, while all configurations tie on IEEE13.

On the raw metrics, adaptive authority lowers the CRE21 bus--step violation
rate from 3.93--7.74\% without filtering to 0.52--3.44\%, and lowers mean
$m_s$ in every case.  This protection has a utility cost relative to no
filter, which is expected because the projection requests additional V2G
support or delays charging.  Across the filtered configurations, mean
departure success remains between 98.93\% and 100\%.  Complete three-run
results are provided in Appendix~\ref{app:secondary}.

Against fixed authority, the adaptive scheduler improves mean reward on all
five CRE21 networks.  It also reduces both voltage metrics on K=8 and K=32.
K=4 and K=16 have similar mean safety under the two budgets, while K=1 moves
along the trade-off by recovering reward at moderately higher voltage risk.
This scale dependence motivates feeder-aware calibration of the authority map.

\textbf{Across-run robustness.}
Across three independently trained policies for every scenario, adaptive
authority reduces both voltage metrics in every individual CRE21 run relative
to no filtering, while departure success remains above 97\% throughout the
filtered runs.  Full run-level statistics are provided in
Appendix~\ref{app:secondary}.

On the secondary IEEE tests, IEEE13 has no voltage violations on the 100-day
evaluation split, so filtering changes nothing.  Relative to No
Filter, adaptive authority reduces both voltage metrics on IEEE33 and IEEE69,
at a substantial reward cost, while retaining 100\% departure success.
Adaptive authority also gives slightly better reward and voltage metrics than
the matched fixed budget on both feeders.  These cross-family checks motivate
feeder-specific threshold calibration.

\subsection{Spatial and Temporal Intervention Patterns}

\begin{figure}[!t]
\centering
\includegraphics[width=\columnwidth]{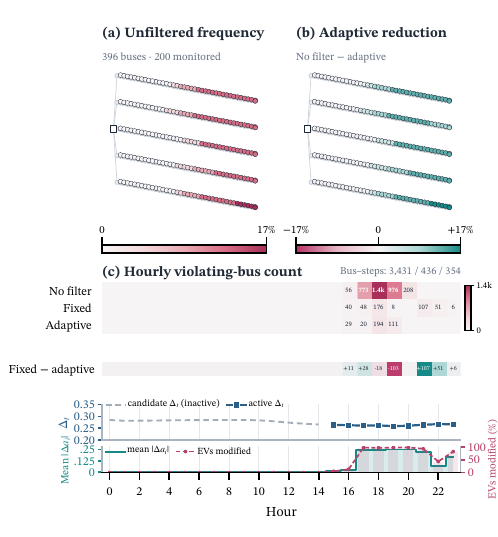}
\caption{Spatial and temporal intervention on a representative K32 evaluation
day.  (a) No Filter violation frequency on the transformer with the largest
unfiltered bus--step count.
(b) Reduction under Adaptive Authority (teal: fewer violating hours).
(c) Hourly violating-bus counts over 3,218 monitored buses, followed by the
Fixed-minus-Adaptive difference, candidate authority (dashed inactive;
squares active), mean pre-service projection correction per EV (teal), and
the share of EVs modified by the projection (rose).  Bus--step totals are
3,431/436/354 for No Filter/Fixed/Adaptive.}
\label{fig:trace}
\end{figure}

Figure~\ref{fig:trace} illustrates how scheduled authority translates into
spatially targeted projection corrections and voltage-risk reduction on K32,
combining topology-resolved exposure, spatial safety gain, and hourly
intervention dynamics.  The representative day is selected as the medoid of
the three arms' daily safety metrics over the evaluation set.

\begin{figure*}[!t]
\centering
\includegraphics[width=0.98\textwidth]{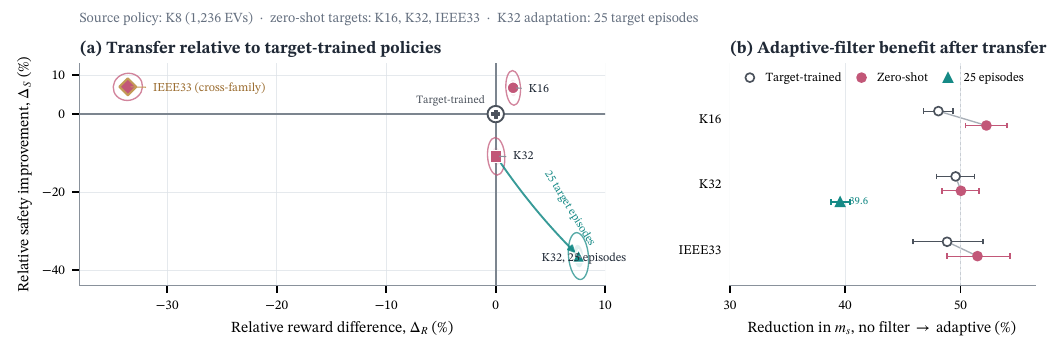}
\caption{Target-normalized topology transfer from the K8 controller.
Panel (a) uses
$\Delta_R=100(R_{\mathrm{tr}}-R_{\mathrm{tgt}})/|R_{\mathrm{tgt}}|$
and
$\Delta_S=100(m_{s,\mathrm{tgt}}-m_{s,\mathrm{tr}})/
m_{s,\mathrm{tgt}}$,
so higher values are preferred and the matched target-trained controller lies
at the origin.  The arrow traces 25-episode K32 adaptation.  Panel (b) reports
the $m_s$ reduction from No Filter to Adaptive Authority.  Ellipses and
whiskers are 95\% paired-day bootstrap regions over the 100 evaluation days
for the matched source checkpoint; they quantify day-level uncertainty.}
\label{fig:transfer}
\end{figure*}

On this day, candidate authority ranges from 0.259 to 0.285 and the filter is
invoked on 9 of 24 steps.  Before activation, both the mean pre-service
projection correction and the fraction of EVs modified by the projection are
zero; across triggered steps, they range from 0.011 to 0.251 and from 4.9\% to
97.7\%, respectively.  Aggregating all 3,218 monitored K32 buses on this day,
Adaptive Authority lowers bus--step violations by 89.7\% relative to No
Filter and by 18.8\% relative to Fixed Authority; the corresponding one-day
$m_s$ values for No Filter, Fixed Authority, and Adaptive Authority are 0.625,
0.461, and 0.289.
The topology panels localize the dominant unfiltered exposure and show that
the corrective benefit is concentrated along the affected radial branches.

\subsection{Scalability and Topology Transfer}

The HetGPS policy-and-risk model has 383,702 learned parameters at deployment;
adding the reward and voltage-cost critics and the auxiliary head used only
during training gives 669,659.  These counts are unchanged from K=1 (200 EVs)
to K=32 (3,218 EVs) because the learned modules are shared; nodewise computation
and nonlearned topology buffers still grow with the graph.  At K=32, the fully
centralized SAC policy has 65,153,316 parameters, about $170\times$ the
deployed HetGPS model.  Adding the two Q-networks used to train that policy
gives 193,799,974 parameters, about $289\times$ the complete HetGPS training
model.  This makes the shared graph controller a compact large-population
architecture.  The transfer experiment below tests whether that shared
parameterization also supports inference on unseen graphs.

The K8 controller transfers cleanly to the larger nested graphs.  With adaptive
authority, zero-shot K16 and K32 achieve violation rates of 0.75\% and 0.57\%
and $m_s$ of 0.324 and 0.338, respectively, close to the corresponding
target-trained operating points (0.94\%/0.347 and 0.51\%/0.305).  Departure
success remains 99.99--100\%.  On IEEE33, zero-shot adaptive authority reduces
violations from 3.57\% to 1.64\%, with a larger reward cost under the shifted
operating regime.  Twenty-five K32 adaptation episodes recover reward from
$-105.38$ to $-97.45$ relative to zero-shot adaptive control, but increase
$m_s$ from 0.338 under zero-shot transfer to 0.415, which is 36.41\% above
the target-trained value of 0.305.

Figure~\ref{fig:transfer} separates transfer fidelity from the filter benefit.
Zero-shot K16 jointly improves normalized reward and safety relative to its
matched target-trained reference, while zero-shot K32 preserves reward within
0.04\%.  The filter nevertheless retains an
$m_s$ reduction of 52.25\%, 50.04\%, and 51.47\% on zero-shot K16, K32, and
IEEE33, respectively.  The IEEE33 point isolates the harder cross-family
utility gap.  On K32, 25 target episodes move the operating point toward
higher reward while trading safety, rather than uniformly improving both.
Nevertheless, for the adapted policy itself, Adaptive Authority reduces $m_s$
from 0.687 without filtering to 0.415, a 39.6\% reduction.

\section{Limitations and Conclusion}

HetGPS addresses a central tension in graph-coupled MARL: protecting shared
constraints without erasing useful learned coordination. Across five CRE21
scales (200--3,218 EVs), Adaptive Authority reduces voltage violations while
preserving departure service; against fixed authority, it improves mean reward
on all five and lowers the mean safety score on four. Shared graph parameters
also enable zero-shot transfer to larger topologies.

Our evidence remains empirical: balanced power flow and $\delta_m=0$ preclude
a pointwise feasibility claim, and certified deployment requires a validated
model-error margin. Yet the learned-authority/physics-direction factorization
extends beyond EV charging to graph-coupled systems with tractable first-order
action-to-constraint sensitivities: physics supplies corrective directions,
while graph learning compresses variable-size network state and assigns
intervention authority. Candidate domains include coordinated inverter/storage
control, district heating, and water networks. Cross-domain validation and
certified margins are important future work.

\clearpage
\appendix
\setcounter{secnumdepth}{2}
These appendices provide the complete mathematical formulation,
implementation and evaluation protocol, full versions of the main-text result
tables, and reproducibility details. All reported controller comparisons use
the operating configuration specified below. Primary learned results report
the mean and sample standard deviation across three independent training runs,
each evaluated on the same 100 reporting days.

\section{Full Constrained Dec-POMDP and Information Assumptions}
\label{app:dec_pomdp}

\subsection{Tuple, global state, and transition}

Let $\mathcal I=\{1,\ldots,N\}$ denote EV agents,
$\mathcal B$ physical buses, and $\mathcal B_{\mathrm{mon}}\subseteq\mathcal B$
the buses monitored by the controller.  We write the finite-horizon
constrained Dec-POMDP as
\begin{equation}
\begin{aligned}
 \mathcal M=\bigl(&\mathcal I,\mathcal S,
 \{\mathcal A_i\}_{i\in\mathcal I},\mathcal P,
 \{\mathcal O_i\}_{i\in\mathcal I},\mathcal Z,r_0,\\
 &\{g_b\}_{b\in\mathcal B_{\mathrm{mon}}},
 p_0,\gamma,\gamma_c,T\bigr).
\end{aligned}
 \label{eq:supp_tuple}
\end{equation}
Here $T$ is the finite decision horizon; all reported episodes use $T=24$
one-hour decisions.
The exogenous process $\xi^t$ contains the real-time electricity price,
household demand, rooftop-PV generation, and EV arrival/departure events.  One
admissible global state is
\begin{equation}
\begin{aligned}
 x^t=\bigl[&\{q_i^t,z_i^t,q_i^{\mathrm{tgt}},t_i^{\mathrm{dep}}\}_{i=1}^{N},
 V^{t-1},\vartheta^{t-1},\xi^t,\zeta^t\bigr],
\end{aligned}
\label{eq:supp_global_state}
\end{equation}
where $q_i^t$ is state of charge (SoC), $z_i^t$ is grid-connection
availability, and $(V^{t-1},\vartheta^{t-1})$ is the latest solved or measured
network state.  The cache $\zeta^t$ carries the finite histories and emitted
boundary values used by the deterministic observation map.  The transition
kernel combines the signed EV power map and SoC
dynamics in Appendix~\ref{app:physical}, balanced AC power flow, and an
exogenous kernel $p_\xi(\xi^{t+1}\mid\xi^t)$.  The experimental simulator
assumes the exogenous kernel is not causally affected by the charging action.
The expectation in every return below is over $p_0$, the exogenous process,
the physical transition, and stochastic policy sampling.

\subsection{Private observation and graph-mediated context}

Let $H$ be the history length and define
\begin{align}
 h_\rho^t&=[\rho^t,\rho^{t-1},\ldots,\rho^{t-H+1}]^\top,\notag\\
 h_{\ell,i}^t&=[\ell_i^{t-H},\ell_i^{t-H+1},\ldots,\ell_i^{t-1}]^\top,\notag\\
 h_{\mathrm{PV},i}^t&=[p_i^{\mathrm{PV},t-H},p_i^{\mathrm{PV},t-H+1},
 \ldots,p_i^{\mathrm{PV},t-1}]^\top.
 \label{eq:supp_history_order}
\end{align}
The price history includes the current price, whereas the load and PV histories
end at the preceding sample.  The private vector available to EV $i$ is
\begin{equation}
\begin{aligned}
 s_i^t=\bigl[&q_i^{\mathrm{obs},t},z_i^{\mathrm{obs},t},\bar t,
 q_i^{\mathrm{tgt}},\bar t_i^{\mathrm{dep}},\\
 &(h_\rho^t)^\top,(h_{\ell,i}^t)^\top,
 (h_{\mathrm{PV},i}^t)^\top\bigr]^\top
 \in\mathbb R^{5+3H},
\end{aligned}
\label{eq:supp_local_obs}
\end{equation}
where $\bar t=t/T$, $\bar t_i^{\mathrm{dep}}=t_i^{\mathrm{dep}}/T$, and
$\rho^t$ is the real-time price.  The normalized time index is needed because a reactive
policy otherwise cannot distinguish equal instantaneous measurements occurring
at different positions in the daily price, load, PV, and mobility cycles.  The
formulation is parameterized by $H$ rather than tied to a numerical input
dimension; the reported implementation uses $H=24$.

The evaluated checkpoints append four deterministic feasibility descriptors,
\begin{equation}
\begin{aligned}
 \chi_i^t&=[s_i^t,\tau_i^t,\delta q_i^t,\rho_i^{\mathrm{ch}},
 \rho_i^{\mathrm{dis}}]^\top\in\mathbb R^{81},\\
 \tau_i^t&=(t_i^{\mathrm{dep}}-t)_+/24,\\
 \delta q_i^t&=(q_i^{\mathrm{tgt}}-q_i^{\mathrm{obs},t})_+,\\
 \rho_i^{\mathrm{ch}}&=P_i^{\mathrm{ch,max}}/(10\,\mathrm{kW}),\\
 \rho_i^{\mathrm{dis}}&=|P_i^{\mathrm{dis,max}}|/(10\,\mathrm{kW}).
\end{aligned}
 \label{eq:supp_checkpoint_features}
\end{equation}
They are deterministic transforms of the current local state and static
charger ratings; they introduce no other household's private state or future
trajectory.  The graph encoder receives operator-side bus measurements and
public feeder parameters defined in Appendix~\ref{app:graph}.  It returns an
electrical context $c_i^t\in\mathbb R^{16}$ for the serving bus of agent $i$.
The evaluated policy observation is
\begin{equation}
 o_i^t=[\chi_i^t\,\|\,c_i^t]\in\mathbb R^{97},\qquad
 a_i^t\sim\pi_i(\cdot\mid o_i^t).
 \label{eq:supp_policy_obs}
\end{equation}
With $\zeta^t$ included in Eq.~\eqref{eq:supp_global_state}, the observation
model is the deterministic map $o_i^t=Z_i(x^t)$ collecting
Eqs.~\eqref{eq:supp_local_obs}--\eqref{eq:supp_policy_obs}.  Its observation
kernel is therefore the Dirac kernel
\begin{equation}
 \mathcal Z_i(o\mid x)=\delta_{Z_i(x)}(o),\qquad
 \mathcal Z=\{\mathcal Z_i\}_{i\in\mathcal I}.
 \label{eq:supp_observation_kernel}
\end{equation}
The policy is reactive because the finite histories are explicit in $s_i^t$
and $\chi_i^t$ is a deterministic current-state augmentation.
Nothing in the Dec-POMDP requires parameter sharing.  The proposed method sets
$\pi_i=\pi_\theta$ for all EVs as a scalable architectural choice, while
different private states and graph contexts still produce different actions.

The information assumptions are therefore:
\begin{itemize}
  \item a charger uses only its own SoC, availability, service target,
  departure time, charger ratings, and local price/load/PV history;
  \item the graph path uses feeder topology and bus-level voltage/power
  measurements available to the network operator;
  \item other households' SoCs and departure schedules are not concatenated
  into an agent observation; and
  \item AC post-action voltages are training labels and evaluation
  measurements, not
  raw-candidate oracle calls made by the deployment filter.
\end{itemize}
Action selection is therefore operator-mediated and information-local: the
operator constructs graph context from feeder-side information, while each
charger combines that context only with its own private state.  This is not
fully local decentralized execution.  It avoids directly exposing one
household's raw EV state to another household, but is not a formal privacy
guarantee.

\subsection{Expected-cost constraints, shaping, and service}

For bus $b$, define the nonnegative voltage-exceedance cost
\begin{equation}
 g_b^t=\pos{V^{\min}-V_b^t}^{2}+\pos{V_b^t-V^{\max}}^{2}.
 \label{eq:supp_bus_cost}
\end{equation}
The formal cooperative objective is
\begin{subequations}
\label{eq:supp_cmdp}
\begin{align}
 \max_{\boldsymbol\pi}\quad
 J_R(\boldsymbol\pi)
 &=\mathbb E_{\tau\sim p_{\boldsymbol\pi}}
 \left[\sum_{t=0}^{T-1}\gamma^t\sum_{i=1}^{N}r_{i,0}^t\right],\\
 \mathrm{s.t.}\quad
 J_{C,b}(\boldsymbol\pi)
 &=\mathbb E_{\tau\sim p_{\boldsymbol\pi}}
 \left[\sum_{t=0}^{T-1}\gamma_c^t g_b^t\right]
 \leq\epsilon_b,
 \quad b\in\mathcal B_{\mathrm{mon}}.
\end{align}
\end{subequations}
This is an expected cumulative-cost constraint, not the assertion
$V_b^t\in[V^{\min},V^{\max}]$ for every stochastic realization.  Charger
ratings, availability masks, and battery-energy bounds are transition
feasibility conditions.  Departure readiness remains a service objective:
late arrival or low arrival SoC can make a pointwise departure target
physically infeasible.

A fixed voltage-related reward signal may be used for learning, and an
adaptive multiplier may act on a voltage-cost critic.  They are not a second
and third copy of a hard voltage constraint.  Appendix~\ref{app:training}
separates the task reward, fixed shaping, cost-critic label, and evaluation
metric.  In particular, the bus--step violation rate and $m_s$ are descriptive
evaluation metrics; they are not optimized again inside
Eq.~\eqref{eq:supp_cmdp}.
Equation~\eqref{eq:supp_cmdp} is the normative cooperative objective.  The
implemented learner uses parameter-shared mean-field reward and voltage-cost
critics, conditioned on each agent's private observation, graph context, and
the fleet mean and standard deviation of raw action proposals, as a scalable
approximation to this team problem.  Mapping each bus cost to its attached
agents yields a parameter-shared mean-field learning approximation; it is not
an exact centralized optimizer of Eq.~\eqref{eq:supp_cmdp} and does not by
itself certify that every expected-cost constraint is met.

\section{Signed EV Dynamics and Balanced AC Power Flow}
\label{app:physical}

\subsection{Experimental signed-power surrogate}

The reported experiments use $\Delta t=1$ hour and a demand-positive signed
simulator power.  With action $a_i^t\in[-1,1]$, availability
$z_i^t\in\{0,1\}$, SoC $q_i^t\in[0,1]$, and
$\eta^{\mathrm{cmd}}=0.90$, define
\begin{equation}
 \bar P_i^t=
 \begin{cases}
 \min\!\left(P_i^{\mathrm{ch,rate}},
 \dfrac{(1-q_i^t)E_i^{\mathrm{cap}}}{\Delta t}\right),
   &a_i^t\geq0,\\
 \min\!\left(P_i^{\mathrm{dis,rate}},
 \dfrac{q_i^tE_i^{\mathrm{cap}}}{\Delta t}\right),
   &a_i^t<0,
 \end{cases}
 \label{eq:supp_action_map}
\end{equation}
and
\begin{equation}
 P_i^{\mathrm{sim},t}=z_i^t\eta^{\mathrm{cmd}}a_i^t\bar P_i^t,
 \qquad
 q_i^{t+1}=q_i^t+
 \frac{P_i^{\mathrm{sim},t}\Delta t}{E_i^{\mathrm{cap}}}.
 \label{eq:supp_soc}
\end{equation}
Writing $\Delta t$ explicitly keeps the power and energy units clear; its
numerical value is one in all reported experiments.  The same
$P_i^{\mathrm{sim},t}$ enters the demand-positive AC load and the SoC
increment.  Thus $\eta^{\mathrm{cmd}}$ is modeled as a symmetric
command-to-power derating shared by every compared method; direction-dependent
conversion losses are outside the present scope.

A loss-aware extension with grid-side power $P_i^{\mathrm{EV},t}$ would instead
use $\eta_i^{\mathrm{ch}}P_i^{\mathrm{EV},t}$ while charging and
$P_i^{\mathrm{EV},t}/\eta_i^{\mathrm{dis}}$ while discharging in the SoC
increment.  Here the two coefficients are one-way efficiencies; if equal,
the corresponding round-trip efficiency is $\eta_i^2$
\cite{shafiekhah_innovative_2018}.

\subsection{Injection convention and network equations}

We use the generator convention: positive $P_b^{\mathrm{inj},t}$ and
$Q_b^{\mathrm{inj},t}$ inject power into the network.  Let
$\mathcal I_b=\{i:b_i=b\}$ be households/EVs attached to bus $b$ and define
$\ell_i^t=P_i^{\mathrm{load},t}+P_i^{\mathrm{sim},t}-P_i^{\mathrm{PV},t}$.
The implemented power-flow update is
\begin{align}
 P_b^{\mathrm{inj},t}
 &=-\sum_{i\in\mathcal I_b}\ell_i^t,
 \label{eq:supp_injection_p}\\
 Q_b^{\mathrm{inj},t}
 &=-\tan(\arccos 0.95)\sum_{i\in\mathcal I_b}[\ell_i^t]_+,
 \label{eq:supp_injection_q}
\end{align}
Thus positive household net demand, including EV power after netting local PV,
is assigned power factor 0.95; a net-exporting household is represented as a
unity-power-factor static generator.  The simulator does not separately assign
reactive power to base load, PV, and the EV.  Charging reduces net injection;
V2G discharge increases it.  This convention is consistent with the balanced AC equations
\begin{align}
 P_b^{\mathrm{inj},t}
 &=V_b^t\sum_{k\in\mathcal B}V_k^t
 \left(G_{bk}\cos\theta_{bk}^t+B_{bk}\sin\theta_{bk}^t\right),\\
 Q_b^{\mathrm{inj},t}
 &=V_b^t\sum_{k\in\mathcal B}V_k^t
 \left(G_{bk}\sin\theta_{bk}^t-B_{bk}\cos\theta_{bk}^t\right),
 \label{eq:supp_acpf}
\end{align}
for $Y=G+\mathrm jB$ and $\theta_{bk}^t=\theta_b^t-\theta_k^t$.
Pandapower computes the AC operating points used for training labels and
evaluation~\cite{thurner_pandapoweropen-source_2018}.

We use the operating band $[0.95,1.05]$ p.u.  The model is a balanced
positive-sequence equivalent: powers are three-phase aggregates, and the
study does not resolve phase imbalance, neutral-conductor effects, or
phase-specific customer voltages.  Line and transformer parameters enter the
power flow, but ampacity and thermal limits are not control constraints.  All
safety statements must therefore be scoped to the reported positive-sequence
voltage metric and monitored buses.

\section{Heterogeneous Graph Features and Architecture}
\label{app:graph}

\subsection{Node and relation types}

The learning graph augments the electrical feeder with EV-agent nodes:
\begin{equation}
 \mathcal G^t=(\mathcal V_{\mathrm{bus}}\cup\mathcal V_{\mathrm{EV}},
 \mathcal E_{\mathrm{elec}}\cup\mathcal E_{\mathrm{attach}}).
 \label{eq:supp_hetero_graph}
\end{equation}
Electrical edges connect bus nodes and represent either a distribution line or
a transformer connection.  Attachment edges connect each EV to exactly one
serving bus; several EVs may share a bus.  Hence only the attachment subgraph
is bipartite between bus and EV nodes.  The full heterogeneous graph is not
itself described as a bipartite electrical network because bus--bus electrical
edges are also present.

\begin{figure}[t]
\centering
\includegraphics[width=\columnwidth]{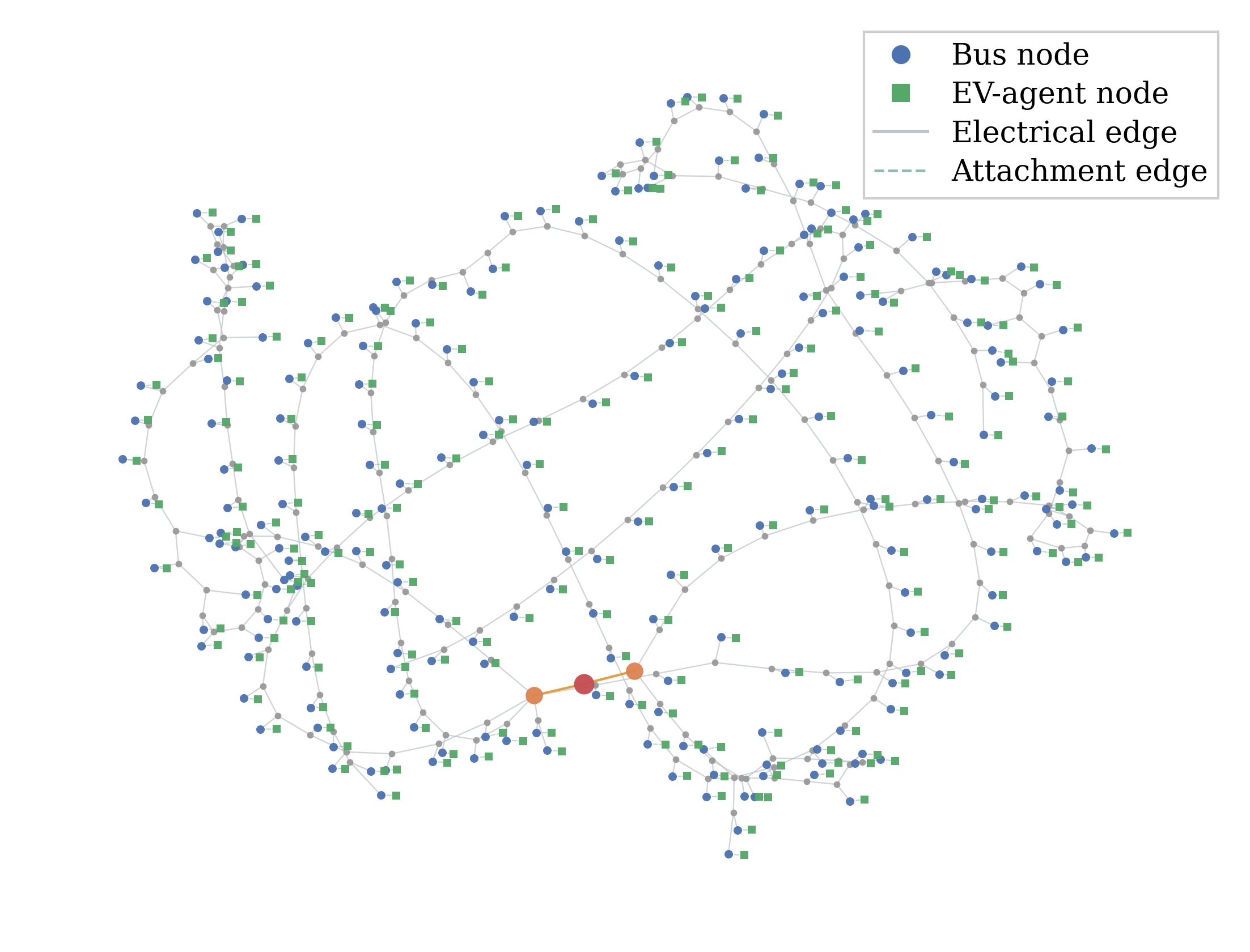}
\caption{Full-network visualization of the heterogeneous graph extracted from
two representative transformer areas of the CRE21 MV/LV network. The learning
graph contains two node types (bus and EV-agent) and two relation types
(electrical and attachment). Bus colours encode physical roles---MV busbar
(red), LV busbar (orange), customer bus (blue), and intermediate bus
(grey)---rather than additional node types. Line and transformer colours are
physical roles within the electrical relation and are distinguished by
$I_{\mathrm{type}}$ in the edge feature. Green squares are EV agents, and
dashed edges are constant-feature attachments. All experiments use the
complete graph of the corresponding scenario.}
\label{fig:supp_heterogeneous_graph}
\end{figure}

Figure~\ref{fig:supp_heterogeneous_graph} visualizes the complete
heterogeneous graph for two representative CRE21 transformer areas.

At decision step $t$, define the fixed-scale net-load and previous-EV features
\begin{align}
 \bar v_{b,t-1}&=\operatorname{clip}\!\left(
 \frac{V_{b,t-1}-1}{0.05},-2,2\right),\\
 \bar p_{b,t}^{\mathrm{net}}&=
 \sum_{i:b_i=b}\frac{P_{i,t}^{\mathrm{load}}-P_{i,t}^{\mathrm{PV}}}
 {10\,\mathrm{kW}},\\
 \bar p_{b,t-1}^{\mathrm{EV}}&=
 \sum_{i:b_i=b}\frac{P_{i,t-1}^{\mathrm{sim}}}
 {P_i^{\mathrm{rate}}}.
 \label{eq:supp_bus_feature_scales}
\end{align}
The exact bus feature interface is
\begin{align}
 x_{b,t}^{\mathrm{bus}}
 &=\left[\bar v_{b,t-1},\bar p_{b,t}^{\mathrm{net}},
 0.33\bar p_{b,t}^{\mathrm{net}},
 \bar p_{b,t-1}^{\mathrm{EV}},\frac{t}{24}\right]^\top,
 \label{eq:supp_bus_features}\\
 x_{i,t}^{\mathrm{EV}}&=\chi_i^t\in\mathbb R^{9+3H},
 \label{eq:supp_ev_features}\\
 e_{bk}^{\mathrm{elec}}&=[L_{bk},R_{bk},X_{bk},I_{bk}^{\mathrm{type}}]^\top,
 \label{eq:supp_edge_features}\\
 e_{i b_i}^{\mathrm{attach}}&=[1]^\top.
 \label{eq:supp_attach_feature}
\end{align}
The second coordinate uses a demand-positive convention, distinct from the
generator convention in Eq.~\eqref{eq:supp_injection_p}.  The third coordinate
is the fixed proxy $0.33\bar p^{\mathrm{net}}$, including its sign under net PV
export; it is not the AC solver's reactive-power injection.  Previous EV power
is normalized agent by agent by rated power before bus aggregation.  For the
slack node (\texttt{MV\_Busbar} or bus \texttt{0}), the voltage coordinate is
set to zero and the three power coordinates are replaced by their mean over all
non-slack buses.  The type indicator is zero for a line and one for a transformer
within the same electrical relation; it does not encode an attachment edge.
Equation~\eqref{eq:supp_attach_feature} separately defines the constant
attachment feature.

Equation~\eqref{eq:supp_checkpoint_features} gives the four feasibility
descriptors used by every reported HetGPS checkpoint.  With $H=24$, the EV
node has 81 coordinates and the actor receives 97 after concatenating its
16-dimensional graph context.

\subsection{Bus encoder, EV readout, and shared actor}

Bus features are projected to $d$ dimensions and updated by $L$ edge-aware
message-passing layers,
\begin{equation}
 h_b^{\ell+1}=\Phi_\ell\!\left(h_b^\ell,
 \{(h_k^\ell,e_{kb}^{\mathrm{elec}}):k\in\mathcal N(b)\}\right).
 \label{eq:supp_bus_mp}
\end{equation}
The large-feeder implementation uses the memory-efficient local-GPS path: the
global token is refreshed by a graph-level mean summary rather than quadratic
all-bus attention.  The reported settings are $d=64$, $L=2$, and four heads in
modules where multi-head attention is active.

For EV $i$, cross-attention reads its serving-bus embedding, selected local
neighbors, and a graph summary using the private vector as query:
\begin{equation}
 h_i^{\mathrm{EV},t}=f_{\mathrm{read}}
 (\chi_i^t,h_{b_i}^t,\{h_b^t:b\in\mathcal N(b_i)\},h_{\mathrm{glob}}^t).
 \label{eq:supp_ev_readout}
\end{equation}
Two learned projections form graph context
$c_i^t=[W_{\mathrm{EV}}h_i^{\mathrm{EV},t}\|W_bh_{b_i}^t]$.
The same Gaussian actor maps $[\chi_i^t\|c_i^t]$ to every agent's action
distribution.  Shared parameters make model size independent of $N$; they do
not constrain the topology size, and transfer is evaluated separately.
Training samples from the tanh-squashed Gaussian policy.  All reported policy
evaluations instead use its deterministic action
$a_i^t=\tanh(\mu_\theta(o_i^t))$, so filter comparisons do not include
additional action-sampling noise.

\subsection{Action-conditioned voltage residual and risk aggregation}

The frozen LinDistFlow matrix is constructed once for the evaluation topology:
\begin{equation}
 [J_{\mathrm{LDF}}]_{bi}=-R^{\mathrm{sens}}_{b,b_i}
 \frac{P_i^{\mathrm{rate}}10^{-3}}{(V_{\mathrm{base}}^{\mathrm{kV}})^2}.
 \label{eq:supp_jldf_build}
\end{equation}
This gives a static, rated-power nominal sensitivity in action coordinates.
For topology transfer, the nonlearned matrix is rebuilt from the target
feeder rather than copied from the source checkpoint.

For a candidate joint action $a$, a shared per-agent action encoder first
embeds each proposal together with its agent representation.  The resulting
embeddings are sum-pooled at the agents' serving buses and processed by the
shared bus residual model; the learned input dimension therefore does not grow
with the fleet.  The residual head predicts
\begin{equation}
 \widetilde V_b^t(a)=V_{b,\mathrm{obs}}^{t-1}
 +[J_{\mathrm{LDF}} a]_b
 +\widehat r_{\psi,b}(\mathcal G^t,a).
 \label{eq:supp_residual_head}
\end{equation}
The residual head is trained from actor proposals and the corresponding
post-action AC voltage labels returned by the environment.  Its pinball loss at
$\tau=0.50$ estimates a conditional median.
The lower and upper heads use $\tau=0.05$ and $0.95$, respectively.  These
auxiliary quantile outputs share the median head's trunk; deployment uses the
median only, and none of the three heads supplies the correction Jacobian.

The reported adaptive scheduler computes risk at the
agent--bus incidences, not at distinct buses.  The unified evaluation uses
internal projection margin $\delta_m=0$ and fixed physics-gain scale
$\beta=0.31$.  Define
\begin{equation}
\begin{aligned}
 z_{b,\psi}^t={}&\pos{V^{\min}+\delta_m-
 \widetilde V_b^t(a^{\mathrm{RL}})}\\[-2pt]
 &+\pos{\widetilde V_b^t(a^{\mathrm{RL}})-V^{\max}+\delta_m},\\
 V_{b,\mathrm{guard}}^t={}&V_{b,\mathrm{obs}}^{t-1}
 +\beta[J_{\mathrm{LDF}}a^{\mathrm{RL}}]_b,\\
 z_{b,\mathrm{guard}}^t={}&\pos{V^{\min}+\delta_m-V_{b,\mathrm{guard}}^t}\\
 &+\pos{V_{b,\mathrm{guard}}^t-V^{\max}+\delta_m},\\
 r_\psi^t={}&Q_{.95}\{z_{b_i,\psi}^t:i\in\mathcal I\},\\
 r_{\mathrm{guard}}^t={}&Q_{.95}\{z_{b_i,\mathrm{guard}}^t:i\in\mathcal I\},\\
 \bar r^t={}&\max\{r_\psi^t,r_{\mathrm{guard}}^t\}.
\end{aligned}
\label{eq:supp_incidence_risk}
\end{equation}
The operator $Q_{.95}$ uses linear quantile interpolation.  Because the set is
indexed by agents, a serving bus is weighted by its attached EV count.  This
weighting is vacuous on CRE21, where each household forms its own bus and the
agent-to-bus map is injective, so the incidence percentile equals the
distinct-bus percentile; it binds only on the IEEE feeders, which attach
10--15 households per customer bus.  The evaluation metrics in
Appendix~\ref{app:secondary} count each distinct EV-hosting bus once
throughout.

The intervention budget is the monotone map
\begin{equation}
 \Delta_t=\Delta_{\min}+(\Delta_{\max}-\Delta_{\min})
 \operatorname{clip}\!\left(
 \frac{\bar r^t-r_{\mathrm{low}}}
 {r_{\mathrm{high}}-r_{\mathrm{low}}},0,1\right).
 \label{eq:supp_authority_map}
\end{equation}
The unified evaluation uses
$(\Delta_{\min},\Delta_{\max})=(0.20,0.35)$ and
$(r_{\mathrm{low}},r_{\mathrm{high}})=(0.005,0.080)$ p.u.  The same
$\beta$-guard identifies the triggered agent--bus incidences, but neither the
learned residual nor $\beta J_{\mathrm{LDF}}$ supplies the correction
Jacobian.  Let
$\widehat V^{\mathrm{phys}}(a^{\mathrm{RL}})=
V_{\mathrm{obs}}^{t-1}+J_{\mathrm{LDF}}a^{\mathrm{RL}}$.
For every triggered lower-side agent--bus incidence at bus $b$, form
$c_j=J_{b,:}^{\mathrm{LDF}}$ and
$h_j=V^{\min}+\delta_m-\widehat V_b^{\mathrm{phys}}$; for an upper-side
incidence, form $c_j=-J_{b,:}^{\mathrm{LDF}}$ and
$h_j=\widehat V_b^{\mathrm{phys}}-(V^{\max}-\delta_m)$.  Repeated agents at
one bus therefore retain the incidence weighting used by the risk aggregation.

The deployed projected solver is a finite sequential-sweep operator.  The
reported experiments use the fixed Euclidean metric $W=I$.  Let $d^{(k)}$
denote the correction after $k$ complete sweeps; constraints update $d$
in-place within each sweep.  Define
\begin{equation}
\begin{aligned}
 \ell_{i,t}&=\max\{-1-a_{i,t}^{\mathrm{RL}},-\Delta_t\},\\
 u_{i,t}&=\min\{1-a_{i,t}^{\mathrm{RL}},\Delta_t\},\\
 d^{(0)}&=0,\\
 g_j(d)&=\frac{[h_j-c_j^\top d]_+}{c_j^\top W^{-1}c_j},\\
 d&\leftarrow\operatorname{proj}_{[\ell_t,u_t]}
 (d+g_j(d)W^{-1}c_j),\\
 a_t^{\mathrm{proj}}&=\operatorname{clip}
 (a_t^{\mathrm{RL}}+d^{(K_{\mathrm{proj}})},-1,1).
\end{aligned}
\label{eq:supp_projection}
\end{equation}
Here $\operatorname{proj}_{[\ell_t,u_t]}$ is componentwise box projection.
Constraints are sorted by initial deficit and
$K_{\mathrm{proj}}=10$ sequential sweeps are used.
When box clipping blocks a weighted update, remaining one-sided authority is
allocated in descending sensitivity-gain order.  The procedure computes the
minimum weighted-norm correction for one unclipped half-space at a time, but is
not an exact joint QP solver and may return a partial action with nonzero
residual.  Availability and SoC are handled by the physical action-to-power
map and subsequent service guard.

The action passed to the voltage filter and the command physically executed are
distinct:
\begin{equation}
 a_t^{\mathrm{raw}}=a_t^{\mathrm{RL}},\qquad
 a_t^{\mathrm{exec}}=G_{\mathrm{svc}}
 (x_t,a_t^{\mathrm{proj}}).
 \label{eq:supp_action_lineage}
\end{equation}
No Filter uses $a_t^{\mathrm{proj}}=a_t^{\mathrm{raw}}$ but retains the same
$G_{\mathrm{svc}}$.  The matched fixed- and adaptive-authority arms share the
same risk forward pass, physics guard, sensitivity model, projection, and
service map.  They differ only in whether the triggered projection uses
Eq.~\eqref{eq:supp_authority_map} or the constant $\Delta_t=0.35$.  The unified
comparison is otherwise unchanged.

\subsection{Directional monotonicity of the physics anchor}

The correction geometry admits a simple sign property under the radial
LinDistFlow approximation used to construct $J_{\mathrm{LDF}}$
\cite{baran_network_1989}.

\begin{proposition}[Voltage-corrective direction]
\label{prop:supp_direction}
Consider a balanced radial feeder with positive line resistances and the
active-power LinDistFlow model used by the filter.  Let
$p_i^{\mathrm{load}}$ denote demand-positive power at the bus $b_i$ serving EV
$i$, and let $v_b$ denote squared voltage magnitude.  Then
\begin{equation}
 \frac{\partial v_b}{\partial p_i^{\mathrm{load}}}
 =-2\!\sum_{e\in\mathcal P(b)\cap\mathcal P(b_i)} r_e\leq0,
 \label{eq:supp_directional_derivative}
\end{equation}
where $\mathcal P(b)$ is the root-to-$b$ path.  Because the implemented EV
power map is nondecreasing in the normalized action, reducing charging or
increasing discharge cannot decrease the LinDistFlow voltage at any bus whose
root path overlaps that of the EV.  The opposite action direction is therefore
appropriate for an overvoltage.
\end{proposition}

\paragraph{Proof.}
For a radial feeder, LinDistFlow gives
$v_b=v_0-2\sum_{e\in\mathcal P(b)}(r_eP_e+x_eQ_e)$.  An incremental active load
at $b_i$ appears in $P_e$ exactly on the edges in $\mathcal P(b_i)$, yielding
Eq.~\eqref{eq:supp_directional_derivative}.  Every term in the shared-path sum
is nonnegative.  Away from action-map saturation,
$\partial p_i^{\mathrm{load}}/\partial a_i\geq0$; at saturation it is zero.
The chain rule therefore gives $\partial v_b/\partial a_i\leq0$.  Thus a
one-sided lower-voltage half-space update takes $d_i\leq0$, while a one-sided
upper-voltage update takes $d_i\geq0$, for agents with nonzero shared-path
sensitivity. $\square$

This proposition establishes the sign of the nominal correction direction.  It
does not assert AC feasibility: finite projection sweeps, nonlinear losses,
the service map, and balanced-model error can still leave residual voltage
violations.  The empirical direction-fidelity diagnostic in
Appendix~\ref{app:direction_diagnostic} compares this first-order direction
directly with AC counterfactual voltage changes.

\section{Reward, SAC, Voltage Cost, and PID Multiplier}
\label{app:training}

\subsection{Task reward and unit conventions}

The primary policy-training loop does not include the deployment voltage
projection.  Its action chain is
\begin{equation}
\begin{aligned}
 a_t^{\mathrm{raw}}&=a_t^{\mathrm{RL}},\\
 a_t^{\mathrm{exec,tr}}&=
 G_{\mathrm{svc}}(x_t,a_t^{\mathrm{raw}}),\\
 \bigl(r_{0}^{t},V_t^{\mathrm{AC}},\widetilde g^t\bigr)
 &\leftarrow
 \operatorname{ACStep}(x_t,a_t^{\mathrm{exec,tr}}).
\end{aligned}
 \label{eq:supp_training_action_lineage}
\end{equation}
Replay stores the raw SAC proposal $a_t^{\mathrm{raw}}$, while the reward and
voltage-cost label are generated after the service map and the ensuing
post-action AC power-flow solve.  Deployment inserts
$a_t^{\mathrm{proj}}$ between $a_t^{\mathrm{raw}}$ and
$G_{\mathrm{svc}}$, as shown in Eq.~\eqref{eq:supp_action_lineage}.

The primary CRE21 runs use the following exact reward decomposition:
\begin{equation}
 r_{i,0}^t=-s_r\left(C_{i,E}^t+C_{i,D}^t+C_{i,S}^t+C_{i,A}^t\right)
 -C_{i,V}^t.
 \label{eq:supp_reward}
\end{equation}
The energy term uses real-time pricing, not a time-of-use tariff:
\begin{equation}
 C_{i,E}^t=10\rho^t
 \left(P_i^{\mathrm{load},t}+P_i^{\mathrm{sim},t}
 -P_i^{\mathrm{PV},t}\right)\Delta t.
 \label{eq:supp_energy_cost}
\end{equation}
A first-order throughput degradation surrogate is
\begin{equation}
 C_{i,D}^t=c^{\mathrm{deg}}|P_i^{\mathrm{sim},t}|\Delta t,
 \label{eq:supp_degradation}
\end{equation}
Here $s_r=0.1$ in Eq.~\eqref{eq:supp_reward}, and
$c^{\mathrm{deg}}=100\times0.00015=0.015$: 100 is the configured battery
replacement-cost scale and 0.00015 is the configured fractional
wear-per-throughput factor.  With the one-hour step, the resulting coefficient
is 0.015 cost units/kWh.  This is a linear training
surrogate, not an electrochemical aging model; throughput surrogates are common
in tractable EV/V2G scheduling
\cite{ortegavazquez_optimal_2014,farzin_practical_2016}.

For a connected EV, let
$d_i^t=\pos{q_i^{\mathrm{tgt}}-q_i^{t+1}}$,
$H_{i,\mathrm{raw}}^t=\max(t_i^{\mathrm{dep}}-t,0)$, and
$H_i^t=\max(H_{i,\mathrm{raw}}^t,1)$.  When
$H_{i,\mathrm{raw}}^t>0$, define
\begin{equation}
 u_i^t=\frac{d_i^tE_i^{\mathrm{cap}}}
 {H_{i,\mathrm{raw}}^tP_i^{\mathrm{ch,rate}}\eta^{\mathrm{cmd}}},\qquad
 \phi(u)=\begin{cases}u,&u\leq1,\\2u-1,&u>1.\end{cases}
\end{equation}
The continuous and departure-event service costs are
\begin{align}
 C_{i,\mathrm{cont}}^t&=
 \begin{cases}
 0,&d_i^t=0,\\
 E_i^{\mathrm{cap}}(d_i^t)^2,
 &d_i^t>0,\ H_{i,\mathrm{raw}}^t\leq0,\\
 0.2E_i^{\mathrm{cap}}d_i^t\phi(u_i^t),
 &d_i^t>0,\ H_{i,\mathrm{raw}}^t>0,
 \end{cases}\\
 C_{i,\mathrm{dep}}^t&=2E_i^{\mathrm{cap}}(d_i^t)^2
 \mathbf1\{t+1=t_i^{\mathrm{dep}}\},\\
 C_{i,S}^t&=\left(10+\frac{10}{0.001+H_i^t}+20d_i^t\right)
 \left(C_{i,\mathrm{cont}}^t+C_{i,\mathrm{dep}}^t\right).
 \label{eq:supp_service_cost}
\end{align}
Thus the implementation does not apply a departure penalty at every step; it
does apply continuous readiness shaping before departure.  This readiness
term is not potential-based shaping; it is part of the optimized task reward.
The remaining terms are
\begin{align}
 C_{i,A}^t&=45(a_i^{\mathrm{reg},t})^2,\\
 C_{i,V}^t&=100\left(\pos{V^{\min}-V_{b_i}^t}^{2}
 +\pos{V_{b_i}^t-V^{\max}}^{2}\right).
 \label{eq:supp_reward_other_terms}
\end{align}
The transformer-overload reward coefficient is zero in these runs.  Here
$a_i^{\mathrm{reg},t}$ is the command seen by the sub-environment after any
connected-EV clamp or emergency override.  For a disconnected EV it remains
the incoming projected command, so the regularizer can be nonzero even though
the physical EV power is zero; it is therefore distinct from the diagnostic
physical-command trace $a_i^{\mathrm{exec},t}$, which zeros disconnected EVs.
For a connected EV, the guard first replaces $a$ by
$\max(a,0)$ only when $H_{i,\mathrm{raw}}^t\leq2$ and
$q_i^t<q_i^{\mathrm{tgt}}$.  It then forces $a=1$ when
$H_{i,\mathrm{raw}}^t>0$ and
\begin{equation}
 d_i^{\mathrm{pre},t}E_i^{\mathrm{cap}}
 \geq0.9H_{i,\mathrm{raw}}^tP_i^{\mathrm{ch,rate}}
 \eta^{\mathrm{cmd}},
 \label{eq:supp_service_guard}
\end{equation}
where $d_i^{\mathrm{pre},t}$ is the deficit before executing the current
command.  Disconnected EV power is zeroed by the physical map.  The guard is
part of the executed action interface and must be included when interpreting
departure success and action-conditioned diagnostic traces.

The fixed voltage shaping signal is $C_{i,V}^t$ in
Eq.~\eqref{eq:supp_reward_other_terms}; it is already included in
$r_{i,0}^t$ and is not subtracted a second time.  Voltage labels are mapped
from serving buses to agents.  When the team loss averages over agents, a bus
with several EVs consequently has several incidences unless explicitly
deduplicated.  This incidence weighting must be distinguished from the
one-bus-one-count evaluation denominator.

\subsection{Twin reward and voltage-cost critics}

For raw fleet proposals, define the mean-field statistic
\begin{equation}
\begin{aligned}
 \bar a^t&=\frac{1}{N}\sum_{k=1}^N a_k^{\mathrm{raw},t},\\
 s_a^t&=\sqrt{\frac{1}{N}\sum_{k=1}^N
 \left(a_k^{\mathrm{raw},t}-\bar a^t\right)^2},\\
 \omega_a^t&=[\bar a^t,s_a^t]^\top .
\end{aligned}
 \label{eq:supp_action_mean_field}
\end{equation}
For replay transition
$(o_i^t,a_i^{\mathrm{raw},t},\omega_a^t,r_{i,0}^t,
\widetilde g_i^t,o_i^{t+1})$, define the
unclipped twin-SAC reward target and its implemented floor by
\begin{align}
 \widetilde y_{R,i}^t={}&r_{i,0}^t+\gamma(1-\mathsf{done}_i^t)
 \Bigl[\min_{j=1,2}\bar Q_{R,j}
 (o_i^{t+1},a_i',\omega_a')\notag\\[-2pt]
 &\hspace{7.5em}{}-\alpha_{\mathrm{ent}}
 \log\pi_\theta(a_i'\mid o_i^{t+1})\Bigr],
 \\
 y_{R,i}^t={}&\max\{-500,\widetilde y_{R,i}^t\},
 \label{eq:supp_reward_target}
\end{align}
with $a_i'\sim\pi_\theta(\cdot\mid o_i^{t+1})$ and $\omega_a'$ formed from
the corresponding fleet of next raw proposals.  The cost target is
\begin{align}
 \widetilde y_{C,i}^t={}&\widetilde g_i^t
 +\gamma(1-\mathsf{done}_i^t)
 \min_{j=1,2}\bar Q_{C,j}(o_i^{t+1},a_i',\omega_a'),\\
 y_{C,i}^t={}&\min\{50,\widetilde y_{C,i}^t\}.
 \label{eq:supp_cost_target}
\end{align}
Thus the reported learner sets the formal cost discount $\gamma_c=\gamma$;
the asymmetric floor/cap are numerical stabilizers.  The critics minimize
squared Bellman errors.  The bus-level cost surrogate used
by the current phase-2 learner is
\begin{equation}
\begin{aligned}
 \widetilde g_b^t={}&100\left(
 \pos{V^{\min}-V_b^{t,\mathrm{AC}}}^{2}
 +\pos{V_b^{t,\mathrm{AC}}-V^{\max}}^{2}\right)\\
 &+10\pos{V^{\min}+0.01-V_b^{t,\mathrm{AC}}}\\
 &+10\pos{V_b^{t,\mathrm{AC}}-(V^{\max}-0.01)}.
\end{aligned}
\label{eq:supp_training_cost}
\end{equation}
The replay label is
$\widetilde g_i^t=\widetilde g_{b_i}^t$, evaluated from the instantaneous
post-action AC voltage produced after the service-mapped training command.
The proximity term supplies signal near the boundary.  Because
Eq.~\eqref{eq:supp_training_cost} is not identical to the formal cost in
Eq.~\eqref{eq:supp_bus_cost}, the multiplier is best described as an adaptive
training penalty for a voltage-cost surrogate, not as a numerical certificate
that Eq.~\eqref{eq:supp_cmdp} has been solved to dual optimality.

For sampled action $a_i\sim\pi_\theta(\cdot\mid o_i)$, the evaluated
\texttt{scale\_all} phase-2 actor loss has the form
\begin{align}
 \mathcal L_\pi&=\mathbb E[(\ell_{R,i}+\ell_{C,i})/s_Q],\\
 \ell_{R,i}&=\alpha_{\mathrm{ent}}\log\pi_\theta(a_i\mid o_i)
 -\min_jQ_{R,j}(o_i,a_i,\omega_a),\\
 \ell_{C,i}&=\beta_e\lambda_{b_i}
 \min_jQ_{C,j}(o_i,a_i,\omega_a),
 \label{eq:supp_actor_loss}
\end{align}
where $s_Q\geq1$ is the moving reward-value scale.  The implementation computes
$\beta_e=\min\{1,e/E_{\mathrm{ann}}\}$, and the evaluated checkpoints are past
the saturation horizon.  Graph features are detached in this actor update; the
graph encoder and residual head receive their voltage-supervision loss.  This
keeps the actor conditioned on the electrical representation while preserving
its supervised voltage objective.

\subsection{Per-bus PID update}

The PID does not average the magnitudes in
Eq.~\eqref{eq:supp_training_cost}.  Its feedback is the per-bus episode event
rate
\begin{equation}
 \widehat c_b^e=\frac{1}{T_e}\sum_{t=0}^{T_e-1}
 \mathbf 1\{\widetilde g_b^{e,t}>0\},
 \label{eq:supp_pid_feedback}
\end{equation}
where $T_e=24$ in the reported runs and
$\widetilde g_b^{e,t}>0$ denotes either a hard voltage violation or entry into
the $0.01$-p.u. proximity band in Eq.~\eqref{eq:supp_training_cost}.  Let
$e_b^e=\widehat c_b^e-c_{\lim}$.  After the configured warm-up, the PID-style
controller is
\begin{align}
 I_b^e&=\operatorname{clip}
 (\nu I_b^{e-1}+K_Ie_b^e,-I_{\max},I_{\max}),\\
 D_b^e&=K_D(e_b^e-e_b^{e-1}),\\
 \lambda_b^{e+1}&=\operatorname{clip}
 (\lambda_b^e+K_Pe_b^e+I_b^e+D_b^e,
 \lambda_{\min},\lambda_{\max}).
 \label{eq:supp_pid}
\end{align}
The reported CRE21 graph runs use
$(K_P,K_I,K_D)=(0.5,0.01,0.05)$, integral decay $\nu=0.99$,
$I_{\max}=1$, $(\lambda_{\min},\lambda_{\max})=(0,3)$,
$c_{\lim}=0.01$, and a 50-episode phase-2 warm-up.  Adaptive cost-limit
calibration is disabled.  The fixed voltage reward term $C_{i,V}^t$ and the
adaptive bus multiplier $\lambda_b$ have different roles.  Neither converts
the deployment filter into an almost-sure safe controller.

\section{Evaluation Metrics and Complete Results}
\label{app:secondary}

\subsection{Metrics and denominator}

Let $\mathcal B_{\mathrm{EV},e}$ be the distinct physical buses serving at
least one EV in episode $e$.  Define
\begin{equation}
 d_{e,t,b}=\pos{V^{\min}-V_{e,t,b}}+
 \pos{V_{e,t,b}-V^{\max}}.
\end{equation}
The aggregate bus--step violation rate is
\begin{equation}
 r_{\mathrm{viol}}=
 \frac{\sum_e\sum_{t=0}^{T_e-1}
 \sum_{b\in\mathcal B_{\mathrm{EV},e}}
 \mathbf1\{d_{e,t,b}>0\}}
 {\sum_eT_e|\mathcal B_{\mathrm{EV},e}|}.
 \label{eq:supp_violation_rate}
\end{equation}
Each serving bus is counted once per hour, irrespective of EV multiplicity.
The per-episode safety score is
\begin{equation}
 m_{s,e}=\frac1{T_e}\sum_{t=0}^{T_e-1}
 \mathbf1\{\max_b d_{e,t,b}>0\}
 +\frac{\max_{t,b}d_{e,t,b}}{V^{\max}-V^{\min}},
 \label{eq:supp_ms}
\end{equation}
and reported $m_s$ is the arithmetic mean over episodes.  Strict exceedance is
used; an exact boundary value is not counted.  These definitions monitor
distinct EV-hosting buses, not every graph node and not agent--bus incidences.
Departure success is counted when the departure SoC plus an absolute tolerance
of 0.10 reaches the requested target; it is not an exact-target indicator.
The 100 reporting episodes are run in scheduler order, with battery SoC carried
from one episode reset to the next after initialization.  Successive days are
therefore statefully linked.  All compared arms use the same ordered days, and
each learned run is first averaged over those days before the mean and sample
standard deviation are computed across the three independent runs.

\subsection{Complete backbone results}

Table~\ref{tab:supp_backbone_results} expands the main-text comparison with
departure-success results.

\begin{table*}[t]
\centering
\scriptsize
\setlength{\tabcolsep}{3pt}
\renewcommand{\arraystretch}{0.88}
\begin{tabular}{llrrrr}
\toprule
Scenario & Method & Reward/EV $\uparrow$ & Viol. (\%) $\downarrow$
& $m_s\downarrow$ & Dep. (\%) $\uparrow$\\
\midrule
CRE21 $K{=}1$ & HetGPS & $\boldsymbol{-100.31\pm2.60}$ & $0.043\pm0.033$ & $0.015\pm0.010$ & $99.85\pm0.05$\\
 & ISAC & $-104.42\pm1.08$ & $\boldsymbol{0.001\pm0.001}$ & $\boldsymbol{0.001\pm0.001}$ & $97.09\pm1.92$\\
 & MADDPG & $-138.90\pm25.36$ & $0.103\pm0.002$ & $0.030\pm0.001$ & $100.00\pm0.01$\\
 & C-SAC & $-353.67\pm17.96$ & $0.000\pm0.000$ & $0.000\pm0.000$ & $79.94\pm1.50$\\
 & PI-LP & $-135.46$ & $1.409$ & $0.133$ & $100.00$\\
\midrule
CRE21 $K{=}4$ & HetGPS & $\boldsymbol{-103.51\pm1.20}$ & $0.058\pm0.012$ & $0.059\pm0.007$ & $100.00\pm0.01$\\
 & ISAC & $-108.38\pm2.11$ & $\boldsymbol{0.017\pm0.002}$ & $\boldsymbol{0.019\pm0.002}$ & $99.47\pm0.77$\\
 & MADDPG & $-115.04\pm4.21$ & $0.133\pm0.002$ & $0.105\pm0.001$ & $100.00\pm0.00$\\
 & C-SAC & $-382.75\pm15.81$ & $0.000\pm0.000$ & $0.000\pm0.000$ & $78.41\pm0.86$\\
 & PI-LP & $-136.31$ & $1.288$ & $0.184$ & $100.00$\\
\midrule
CRE21 $K{=}8$ & HetGPS & $\boldsymbol{-103.62\pm2.63}$ & $0.019\pm0.010$ & $0.033\pm0.014$ & $99.89\pm0.19$\\
 & ISAC & $-121.89\pm3.14$ & $\boldsymbol{0.002\pm0.002}$ & $\boldsymbol{0.005\pm0.005}$ & $100.00\pm0.00$\\
 & MADDPG & $-126.76\pm23.53$ & $0.083\pm0.005$ & $0.096\pm0.004$ & $99.79\pm0.36$\\
 & C-SAC & $-383.91\pm2.29$ & $0.000\pm0.000$ & $0.000\pm0.000$ & $78.06\pm0.34$\\
 & PI-LP & $-136.88$ & $1.171$ & $0.183$ & $100.00$\\
\midrule
CRE21 $K{=}16$ & HetGPS & $\boldsymbol{-107.22\pm5.43}$ & $0.007\pm0.012$ & $0.020\pm0.032$ & $99.87\pm0.22$\\
 & ISAC & $-127.33\pm6.49$ & $\boldsymbol{0.005\pm0.006}$ & $\boldsymbol{0.016\pm0.020}$ & $100.00\pm0.00$\\
 & MADDPG & $-136.31\pm30.35$ & $0.072\pm0.020$ & $0.109\pm0.015$ & $99.99\pm0.02$\\
 & C-SAC & $-389.18\pm7.60$ & $0.000\pm0.000$ & $0.000\pm0.000$ & $77.60\pm0.87$\\
 & PI-LP & $-136.85$ & $1.137$ & $0.204$ & $100.00$\\
\midrule
CRE21 $K{=}32$ & HetGPS & $\boldsymbol{-108.45\pm8.21}$ & $\boldsymbol{0.001\pm0.001}$ & $\boldsymbol{0.003\pm0.003}$ & $99.89\pm0.12$\\
 & ISAC & $-126.78\pm4.97$ & $0.002\pm0.002$ & $0.012\pm0.014$ & $100.00\pm0.00$\\
 & MADDPG & $-153.99\pm0.02$ & $0.061\pm0.001$ & $0.122\pm0.001$ & $100.00\pm0.00$\\
 & C-SAC & $-389.56\pm4.51$ & $0.000\pm0.000$ & $0.000\pm0.000$ & $77.92\pm0.75$\\
 & PI-LP & $-137.69$ & $1.010$ & $0.201$ & $100.00$\\
\midrule
IEEE13 & HetGPS & $-190.65\pm0.01$ & $\boldsymbol{0.000\pm0.000}$ & $\boldsymbol{0.000\pm0.000}$ & $100.00\pm0.00$\\
 & ISAC & $-213.91\pm5.11$ & $\boldsymbol{0.000\pm0.000}$ & $\boldsymbol{0.000\pm0.000}$ & $100.00\pm0.00$\\
 & MADDPG & $\boldsymbol{-189.80\pm0.01}$ & $\boldsymbol{0.000\pm0.000}$ & $\boldsymbol{0.000\pm0.000}$ & $100.00\pm0.00$\\
 & C-SAC & $-1874.91\pm128.20$ & $0.000\pm0.000$ & $0.000\pm0.000$ & $74.46\pm2.09$\\
 & PI-LP & $-592.47$ & $0.125$ & $0.017$ & $100.00$\\
\midrule
IEEE33 & HetGPS & $\boldsymbol{-191.29\pm0.07}$ & $\boldsymbol{3.529\pm0.001}$ & $0.441\pm0.001$ & $100.00\pm0.00$\\
 & ISAC & $-214.20\pm2.14$ & $3.646\pm0.019$ & $0.425\pm0.003$ & $100.00\pm0.00$\\
 & MADDPG & $-192.06\pm0.04$ & $3.650\pm0.001$ & $0.441\pm0.001$ & $100.00\pm0.00$\\
 & C-SAC & $-2014.02\pm75.71$ & $0.227\pm0.063$ & $0.039\pm0.011$ & $72.23\pm1.37$\\
 & PI-LP & $-479.94$ & $3.814$ & $\boldsymbol{0.336}$ & $100.00$\\
\midrule
IEEE69 & HetGPS & $\boldsymbol{-188.75\pm0.01}$ & $\boldsymbol{0.320\pm0.001}$ & $0.065\pm0.001$ & $100.00\pm0.00$\\
 & ISAC & $-219.68\pm2.81$ & $0.409\pm0.003$ & $\boldsymbol{0.050\pm0.001}$ & $100.00\pm0.00$\\
 & MADDPG & $-189.63\pm0.07$ & $0.410\pm0.000$ & $0.065\pm0.001$ & $100.00\pm0.00$\\
 & C-SAC & $-2029.39\pm43.36$ & $0.000\pm0.000$ & $0.000\pm0.000$ & $72.66\pm0.26$\\
 & PI-LP & $-570.14$ & $2.101$ & $0.197$ & $100.00$\\
\bottomrule
\end{tabular}
\caption{Complete backbone and PI-LP comparison.  Each learned run is first
averaged over the same 100 reporting days; entries then report mean $\pm$
sample standard deviation across three training runs.  PI-LP is deterministic.
Bold marks the best reward or safety value among controllers with at least
95\% departure success.}
\label{tab:supp_backbone_results}
\end{table*}

\subsection{Detailed authority-filter results}

Table~\ref{tab:supp_authority_results} reports the exact operating points used
by the main reward--safety analysis.

\begin{table*}[t]
\centering
\scriptsize
\setlength{\tabcolsep}{3.5pt}
\begin{tabular}{llrrrr}
\toprule
Scenario & Configuration & Reward/EV & Viol. (\%) & $m_s$ & Dep. (\%)\\
\midrule
CRE21 $K=1$  & No Filter          & $-85.58\pm2.73$  & $7.739\pm0.687$ & $0.591\pm0.048$ & $99.80\pm0.06$ \\
CRE21 $K=1$  & Fixed Authority    & $-99.46\pm3.67$  & $2.676\pm1.144$ & $0.253\pm0.063$ & $99.20\pm0.11$ \\
CRE21 $K=1$  & Adaptive Authority & $-95.57\pm3.03$  & $3.438\pm1.194$ & $0.281\pm0.062$ & $99.35\pm0.11$ \\
\midrule
CRE21 $K=4$  & No Filter          & $-88.25\pm1.25$  & $6.945\pm0.216$ & $0.710\pm0.018$ & $100.00\pm0.00$ \\
CRE21 $K=4$  & Fixed Authority    & $-101.78\pm0.29$ & $1.937\pm0.264$ & $0.375\pm0.013$ & $99.92\pm0.07$ \\
CRE21 $K=4$  & Adaptive Authority & $-101.18\pm0.49$ & $1.949\pm0.302$ & $0.372\pm0.016$ & $99.93\pm0.08$ \\
\midrule
CRE21 $K=8$  & No Filter          & $-87.70\pm2.52$  & $6.383\pm0.261$ & $0.670\pm0.017$ & $99.80\pm0.34$ \\
CRE21 $K=8$  & Fixed Authority    & $-101.11\pm1.58$ & $1.513\pm0.263$ & $0.339\pm0.008$ & $99.10\pm1.44$ \\
CRE21 $K=8$  & Adaptive Authority & $-101.09\pm1.58$ & $1.498\pm0.280$ & $0.332\pm0.016$ & $99.10\pm1.44$ \\
\midrule
CRE21 $K=16$ & No Filter          & $-91.10\pm8.19$  & $5.027\pm0.497$ & $0.639\pm0.051$ & $99.78\pm0.38$ \\
CRE21 $K=16$ & Fixed Authority    & $-109.25\pm3.83$ & $0.798\pm0.773$ & $0.354\pm0.059$ & $99.06\pm1.62$ \\
CRE21 $K=16$ & Adaptive Authority & $-106.32\pm2.89$ & $0.885\pm0.710$ & $0.346\pm0.067$ & $99.06\pm1.62$ \\
\midrule
CRE21 $K=32$ & No Filter          & $-89.16\pm5.32$  & $3.926\pm0.918$ & $0.554\pm0.092$ & $99.73\pm0.26$ \\
CRE21 $K=32$ & Fixed Authority    & $-109.81\pm6.59$ & $0.980\pm0.956$ & $0.428\pm0.085$ & $98.93\pm1.08$ \\
CRE21 $K=32$ & Adaptive Authority & $-107.87\pm6.93$ & $0.516\pm0.225$ & $0.364\pm0.056$ & $99.12\pm0.85$ \\
\midrule
IEEE13 & No Filter          & $-190.50\pm0.01$ & $0.000\pm0.000$ & $0.000\pm0.000$ & $100.00\pm0.00$ \\
IEEE13 & Fixed Authority    & $-190.50\pm0.01$ & $0.000\pm0.000$ & $0.000\pm0.000$ & $100.00\pm0.00$ \\
IEEE13 & Adaptive Authority & $-190.50\pm0.01$ & $0.000\pm0.000$ & $0.000\pm0.000$ & $100.00\pm0.00$ \\
\midrule
IEEE33 & No Filter          & $-192.07\pm0.07$ & $3.604\pm0.001$ & $0.430\pm0.001$ & $100.00\pm0.00$ \\
IEEE33 & Fixed Authority    & $-298.32\pm0.13$ & $1.888\pm0.003$ & $0.223\pm0.001$ & $100.00\pm0.00$ \\
IEEE33 & Adaptive Authority & $-293.74\pm1.22$ & $1.868\pm0.004$ & $0.220\pm0.001$ & $100.00\pm0.00$ \\
\midrule
IEEE69 & No Filter          & $-189.67\pm0.07$  & $0.414\pm0.000$ & $0.065\pm0.001$ & $100.00\pm0.00$ \\
IEEE69 & Fixed Authority    & $-260.36\pm0.09$  & $0.028\pm0.001$ & $0.004\pm0.001$ & $100.00\pm0.00$ \\
IEEE69 & Adaptive Authority & $-253.37\pm10.29$ & $0.014\pm0.013$ & $0.002\pm0.002$ & $100.00\pm0.00$ \\
\bottomrule
\end{tabular}
\caption{Complete matched authority-filter results.  Each run is first
averaged over the same 100 reporting days; entries then report mean $\pm$
sample standard deviation across three independently trained runs.  The three
filter arms use the same trained policy within each run.}
\label{tab:supp_authority_results}
\end{table*}

The backbone and authority comparisons use the same feeder scales, exogenous
profiles, EV fleet, action and service maps, and 100-day reporting set.  They
answer different controlled questions and use their respective three-run
policy cohorts; consequently, the HetGPS backbone row and the No Filter
authority row are not the same checkpoint realization.  Within every
authority-comparison run, however, the policy is held fixed and only the
deployment filter changes.

\subsection{Learned-Direction Ablations}
\label{app:learned_direction}

The learned-authority/physical-direction split was selected empirically rather
than imposed a priori.  We also tested a learned-surrogate-gradient path in
which the graph voltage model selected both correction direction and magnitude,
and a residual-Jacobian path in which the learned model augmented the physical
sensitivity,
\begin{equation}
 \widehat J_\psi
 =J_{\mathrm{LDF}}+\frac{\partial\widehat r_\psi}{\partial a}.
 \label{eq:supp_learned_jacobian}
\end{equation}
The comparison matches the action box and per-action cap across the two
projection implementations so that directional quality is not confounded by a
larger intervention budget.

The directional audit used finite-difference AC power-flow derivatives for 480
sampled action columns at each of three line-impedance multipliers.  The median
cosine similarity with the AC direction was $0.996$--$0.999$ for
$J_{\mathrm{LDF}}$, compared with $0.775$--$0.988$ for $\widehat J_\psi$.
The corresponding median column errors were $0.00440$--$0.04366$ and
$0.00714$--$0.04350$, respectively.  Thus, the residual Jacobian did not
consistently improve either directional alignment or magnitude error.

The paired closed-loop diagnostic below uses the same CRE21 K1 operating
configuration and 100 reporting days.  At each cap, both paths share the policy
checkpoint, day order, action box, and maximum per-action deviation.
\begin{table}[t]
\centering
\small
\setlength{\tabcolsep}{3.4pt}
\begin{tabular}{crrrr}
\toprule
Cap & \multicolumn{2}{c}{Reward/EV $\uparrow$}
& \multicolumn{2}{c}{Viol. (\%) $\downarrow$}\\
\cmidrule(lr){2-3}\cmidrule(lr){4-5}
& Physics & Learned & Physics & Learned\\
\midrule
$0.2$ & $-98.43$  & $-85.26$  & $3.81$ & $6.78$\\
$0.3$ & $-113.66$ & $-86.16$  & $1.58$ & $7.10$\\
$0.5$ & $-151.92$ & $-116.32$ & $0.01$ & $0.28$\\
\bottomrule
\end{tabular}
\caption{Equal-cap learned-direction diagnostic.  ``Physics'' uses
$J_{\mathrm{LDF}}$; ``Learned'' uses $\widehat J_\psi$.}
\label{tab:supp_direction_closed_loop}
\end{table}
The learned direction preserves more reward but leaves more violations at all
three caps, so it does not give a consistent reward--safety advantage over the
physics direction.  Together with the AC derivative audit, this supports using
graph learning to schedule intervention authority while retaining feeder
physics for the correction direction.

\subsection{Per-scenario feeder configuration}

Table~\ref{tab:supp_scenario_config} lists the feeder settings behind every
evaluated scenario.  All scales multiply the corresponding quantity of the
published network, so a value of $1.0$ leaves it untouched.  Full EV
participation, rather than any of these scales, is what drives the voltage
excursions: each household hosts an EV whose charging power exceeds its
residential base load by roughly an order of magnitude.  The scales retune the
feeder around that fleet.  Raising transformer capacity keeps thermal limits
from binding before voltage; reducing base load and line impedance on the
nested CRE21 feeders keeps violations material without making them
unrecoverable, since those long low-voltage cables would otherwise collapse
under full participation.  The two smaller IEEE feeders are evaluated
unmodified, and the elevated PV scale on the IEEE networks admits reverse-flow
overvoltage in addition to undervoltage.

\begin{table}[t]
\centering
\small
\begin{tabular}{lrrrr}
\toprule
Scenario & Tx cap. & PV & Load & Impedance\\
\midrule
CRE21 K1--K32          & 4.0  & 1.0 & 0.7 & 0.3\\
IEEE13, IEEE33         & 1.0  & 3.0 & 1.0 & 1.0\\
IEEE69                 & 10.0 & 3.0 & 0.8 & 0.4\\
\bottomrule
\end{tabular}
\caption{Feeder scales used throughout the backbone, authority-filter, and
transfer experiments, as multipliers of the published networks.  EV
participation is full in every scenario.  IEEE feeders draw 10--15 households
per customer bus with assignment seed 42.}
\label{tab:supp_scenario_config}
\end{table}

The optimization reference is a perfect-information, V2G-enabled PI-LP with a
linearized voltage model and soft voltage slack.  It enforces its own SoC and
departure constraints but is neither an AC-feasible oracle nor a reward upper
bound.  Across CRE21 K1--K32 its reward/EV ranges from $-137.69$ to $-135.46$,
bus--step violation rate from 1.010\% to 1.409\%, and $m_s$ from 0.133 to
0.204, with 100\% departure success.  Because it uses full-day future
profiles, it is an optimization feasibility reference rather than an
information-matched online baseline.

\subsection{Parameter and wall-clock accounting}

\begin{table}[t]
\centering
\small
\setlength{\tabcolsep}{3pt}
\begin{tabular}{lrr}
\toprule
Learned path & Parameters K1 & Parameters K32\\
\midrule
Proposed: deployed & 383,702 & 383,702\\
Proposed: training & 669,659 & 669,659\\
Centralized: actor & 4,111,248 & 65,153,316\\
Centralized: training & 12,231,058 & 193,799,974\\
\bottomrule
\end{tabular}
\caption{Matched parameter accounting at 200 and 3218 EVs.  Proposed
deployment counts the graph encoder/projections, actor, and active residual
risk head; the training row additionally counts reward/cost critics and the
trained counterfactual auxiliary head.  Centralized rows use the corresponding
actor and actor-plus-twin-critic scopes.  Frozen targets, SAC temperature, PID
state, and nonlearned topology buffers are excluded.}
\label{tab:supp_params}
\end{table}

This table compares learned parameters, not total deployment memory.  At K=32,
the nonlearned dense LinDistFlow sensitivity has shape
$6375\times3218$ (20,514,750 entries), approximately 82 MB as float32, and its
storage scales with the product of buses and agents.  Sparse or
feeder-local sensitivities are therefore a natural direction for scaling beyond
the reported network.

The reported simulated timing measurements show 13--390 ms additional
wall-clock per hourly decision for fixed-authority filtering across the eight
CRE21/IEEE scenarios (3.7--8.6\% relative to the corresponding no-filter
evaluation loop).  The isolated percentile/guard/authority rule costs
only aggregation and a scalar map after the common trigger; no separate
micro-timing result is reported for this scalar rule.
Timing is descriptive of the recorded software/hardware stack, not an
algorithmic complexity theorem.  The fixed and adaptive configurations use the
same residual output and projection; adaptive authority adds only incidence-risk
aggregation and a scalar map, but no second residual-head evaluation or
raw-candidate pandapower solve.

\section{Reproducibility Details}
\label{app:reproducibility}

\subsection{Software, data, and deterministic assignments}

The reported implementation uses Python 3.10.14, PyTorch 2.1.1+cu121,
PyTorch Geometric 2.4.0,
pandapower 2.14.10, NumPy 1.26.4, SciPy 1.13.1, CVXPY 1.7.3, and CasADi 3.7.2.
Every episode has 24 one-hour
steps.  Residential load/PV profiles, AEMO real-time prices, and VISTA-style
trip records follow \citet{ratnam_residential_2017},
\citet{aemo_price_demand_2022}, and \citet{vista_2020}.  The heterogeneous EV
fleet is generated once with seed 2024: 40 kWh/7.4 kW (30\%),
58 kWh/11 kW (40\%), and 77 kWh/22 kW (30\%).  IEEE household assignment uses
seed 42.  The day scheduler allocates 240 of 365 days to source-policy
learning, 25 to the K32 few-shot topology-transfer adaptation block, and 100
to final evaluation.  In a 100-day evaluation rollout, resets advance the
scheduled day and mobility variables while retaining the preceding battery
SoC after the first episode.
\subsection{Protocol records}

For every reported comparison, the experiment record includes:
\begin{enumerate}
  \item the exact command, code revision, checkpoint hash, topology list,
  data split, training seed, fleet seed, and household-assignment seed;
  \item architecture compatibility between training and evaluation
  (embedding 64, four heads, two bus layers, shared parameters, simplified
  bus graph, MV busbar, GNN-Lite, and the global-attention setting);
  \item the feature-mode input dimension and the action/SoC invariant tests
  identified in Appendices~\ref{app:physical} and~\ref{app:graph};
  \item the denominator definition based on distinct monitored EV-hosting buses for
  Eq.~\eqref{eq:supp_violation_rate}, including the count for each topology;
  \item the exact reward decomposition and units, including continuous service
  shaping, action regularization, and degradation coefficient;
  \item three aligned 100-day evaluation vectors, one per independent run,
  across no-filter, fixed-authority, and adaptive-authority modes;
  and
  \item the compute environment and anonymized reporting metadata.
\end{enumerate}

The experimental network is a balanced model and the online simulation uses
cached previous-step voltage feedback.  A real deployment would require a
measurement/state-estimation interface and would not query pandapower for the
raw candidate action. 

\section{Topology Transfer and Physics Diagnostics}
\label{app:transfer}

\subsection{Zero-shot and few-shot protocol}

The transfer source is the K8 source checkpoint trained with 1236 EVs.
Zero-shot loading copies learned tensors but preserves target-topology buffers:
the target electrical edge index/attributes, attachment map, household-to-bus
map, and LinDistFlow Jacobian are constructed from the target feeder and are
not overwritten by the source checkpoint.  A successful state-dictionary load
is necessary but not sufficient evidence of transfer.

The fixed evaluation targets are:
\begin{itemize}
  \item K8$\rightarrow$K16 and K8$\rightarrow$K32 under the source-matched
  CRE21 source operating regime, isolating nested topology/fleet scale;
  \item K8$\rightarrow$IEEE33 under the IEEE target regime, an intentionally
  harder topology plus operating-distribution shift; and
  \item K8$\rightarrow$K32 adaptation for exactly 25 target training episodes,
  initialized from the transferable weights of the same source checkpoint,
  with fresh optimizer and replay state and reset target-topology dual state.
\end{itemize}
For each target, evaluate no filter and adaptive authority on the identical
100 held-out evaluation days.  Only the K32 few-shot arm updates its weights
on the separate 25-day adaptation block; these days do not enter the reported
evaluation means.
Report reward/EV, bus--step violation rate, $m_s$, departure success, strict
transferable-tensor loading success with target buffers preserved, and wall-clock.
The target-trained checkpoint is a reference,
not part of zero-shot adaptation.  Few-shot budgets count target environment
episodes, not gradient steps.  During the 25 target episodes, periodic
evaluation and validation-based model selection are disabled; the
predetermined final iterate is evaluated.  The reported intervals summarize
paired-day variation for one source initialization.

In Table~\ref{tab:supp_transfer_results}, ZS denotes zero-shot transfer, 25E
denotes 25 target-training episodes, and NF/AA denote No Filter and Adaptive
Authority, respectively.
\begin{table}[t]
\centering
\scriptsize
\setlength{\tabcolsep}{1.8pt}
\begin{tabular}{llcrrrr}
\toprule
Target & Adapt. & Filter & R/EV & Viol. & $m_s$ & Dep.\\
\midrule
K16    & ZS  & NF & $-90.57$  & 5.80 & .678 & 100.00\\
K16    & ZS  & AA & $-106.29$ & .75  & .324 & 99.99\\
K32    & ZS  & NF & $-90.78$  & 5.17 & .676 & 100.00\\
K32    & ZS  & AA & $-105.38$ & .57  & .338 & 100.00\\
K32    & 25E & NF & $-91.41$  & 5.38 & .687 & 100.00\\
K32    & 25E & AA & $-97.45$  & 2.04 & .415 & 100.00\\
IEEE33 & ZS  & NF & $-289.25$ & 3.57 & .422 & 100.00\\
IEEE33 & ZS  & AA & $-392.89$ & 1.64 & .205 & 100.00\\
\bottomrule
\end{tabular}
\caption{K8-source topology transfer over 100 reporting days. Reward is per
EV; violation and departure values are percentages.}
\label{tab:supp_transfer_results}
\end{table}

Zero-shot K16 and K32 closely track the corresponding target-trained adaptive
operating points: violation/$m_s$ are .75\%/.324 versus .94\%/.347 on K16 and
.57\%/.338 versus .51\%/.305 on K32.  The cross-family IEEE33 transfer remains
safe but incurs a larger reward penalty.  On K32, 25 adaptation episodes move
the policy toward higher reward and a less conservative safety operating point.

\subsection{AC validation of the physical direction}
\label{app:direction_diagnostic}

The direction-fidelity analysis compares the voltage change predicted by the
nominal physical anchor with the AC counterfactual change caused by the same EV
actions.  For each evaluated hour, define
\begin{equation}
\begin{aligned}
 \Delta V_t^{\mathrm{LDF}}&=J_{\mathrm{LDF}}a_t^{\mathrm{exec}},\\
 \Delta V_t^{\mathrm{AC}}&=V_t^{\mathrm{AC}}(a_t^{\mathrm{exec}})
 -V_t^{\mathrm{AC}}(0),\\
 s_t&=\frac{(\Delta V_t^{\mathrm{LDF}})^\top
 \Delta V_t^{\mathrm{AC}}}
 {\|\Delta V_t^{\mathrm{LDF}}\|_2
  \|\Delta V_t^{\mathrm{AC}}\|_2},\\
 \gamma_t&=\frac{(\Delta V_t^{\mathrm{LDF}})^\top
 \Delta V_t^{\mathrm{AC}}}
 {\|\Delta V_t^{\mathrm{LDF}}\|_2^2}.
\end{aligned}
\label{eq:supp_direction_metrics}
\end{equation}
Here $s_t$ measures directional alignment and $\gamma_t$ is the least-squares
AC-to-LinDistFlow gain along the nominal direction.  The zero-EV replay keeps
the same non-EV load, PV, and network state.  This is a diagnostic of correction
geometry, not an additional controller arm or an AC-feasibility certificate.
It uses 2,400 K8 hourly samples from the 100 reporting days and 408 K32
samples from a 17-day diagnostic subset.

\begin{figure}[t]
\centering
\includegraphics[width=\columnwidth]{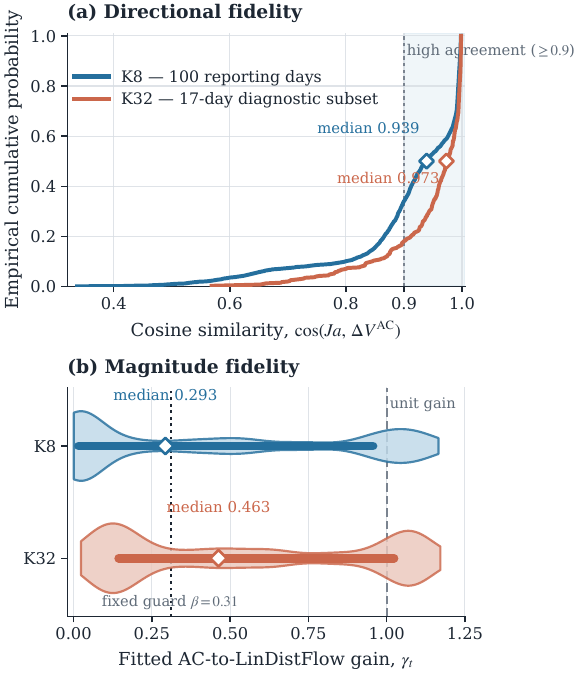}
\caption{Physics-anchor validation on K8 and K32. Directional alignment is
positive throughout (median $s_t=0.939$ and $0.973$), whereas fitted magnitude
gain is feeder dependent (median $\gamma_t=0.293$ and $0.463$), supporting
physics-directed correction with learned authority scheduling.}
\label{fig:supp_direction_fidelity}
\end{figure}

\clearpage
\bibliography{references}

\end{document}